\documentclass[final,5p,times,twocolumn]{elsarticle}

\usepackage{amssymb}
\usepackage{graphicx}
\usepackage[cmex10]{amsmath}
\usepackage{color}
\usepackage{multirow}
\usepackage{makecell}
\usepackage[linesnumbered,ruled]{algorithm2e}
\usepackage{mathrsfs}
\usepackage{picins}
\biboptions{numbers,sort&compress}
\definecolor{grey}{gray}{0.4}
\definecolor{lightorange}{RGB}{255,191,170}

\usepackage[tight,footnotesize]{subfigure}
\usepackage{overpic}
\usepackage{bm}
\usepackage{makecell}
\usepackage{ragged2e}
\usepackage{textcomp}
\usepackage{eso-pic}

\journal{Neurocomputing}

\begin{document}

\begin{frontmatter}

\title{Learning Hough Regression Models via Bridge Partial Least Squares for Object Detection}

\author[a1,a2,a3,a4]{Jianyu Tang}
\ead{jianyu.tang@ieee.org}
\author[a1]{Hanzi Wang\corref{corr}}
\ead{hanzi.wang@ieee.org}
\author[a1]{Yan Yan}
\ead{yanyan@xmu.edu.cn}

\cortext[corr]{Corresponding author. Tel.:+86-592-2580063}

\address[a1]{School of Information Science and Technology, Xiamen University, Xiamen 361005, China}
\address[a2]{Fujian Key Laboratory of the Brain-like Intelligent Systems (Xiamen University), Xiamen 361005, China}
\address[a3]{Cognitive Science Department, Xiamen University, Xiamen 361005, China}
\address[a4]{School of Information Management, Hubei University of Economics, Hubei 430205, China}

\begin{abstract}
Popular Hough Transform-based object detection approaches usually construct an appearance codebook by clustering local image features. However, how to choose appropriate values for the parameters used in the clustering step remains an open problem. Moreover, some popular histogram features extracted from overlapping image blocks may cause a high degree of redundancy and multicollinearity. In this paper, 
we propose a novel Hough Transform-based object detection approach. First, to address the above issues, we exploit a Bridge Partial Least Squares (BPLS) technique to establish context-encoded Hough Regression Models (HRMs), which are linear regression models that cast probabilistic Hough votes to predict object locations. BPLS is an efficient variant of Partial Least Squares (PLS). PLS-based regression techniques (including BPLS) can reduce the redundancy and eliminate the multicollinearity of a feature set. And the appropriate value of the only parameter used in PLS (i.e., the number of latent components) can be determined by using a cross-validation procedure. Second, to efficiently handle object scale changes, we propose a novel \textit{multi-scale voting scheme}. In this scheme, multiple Hough images corresponding to multiple object scales can be obtained simultaneously. Third, an object in a test image may correspond to multiple true and false positive hypotheses at different scales. Based on the proposed multi-scale voting scheme, a principled strategy is proposed to fuse hypotheses to reduce false positives by evaluating \textit{normalized pointwise mutual information} between hypotheses. {In the experiments, we also compare the proposed HRM approach with its several variants to evaluate the influences of its components on its performance.} Experimental results show that the proposed HRM approach has achieved desirable performances on popular benchmark datasets.
\end{abstract}


\end{frontmatter}

\section{Introduction} \label{Intro}
The basic idea of most Hough Transform-based object detection approaches \cite{ISM08,HF09,MMHT,MultiHT,partISM,TUDMotor,FastPRISM} is to model the relationship between local image features and voting points by training a codebook of local appearance. All image features in a test image are extracted and mapped to a number of voting points by using the codebook. All the voting points form a Hough image. The positions of the local maxima in a Hough image are considered to be the locations of object hypotheses. 

A codebook of local appearance is usually constructed by using a clustering approach. For example, the Implicit Shape Model (ISM) \cite{ISM08} employs an agglomerative clustering approach to cluster local image features. The obtained cluster centers form a codebook. A distance threshold is used during the clustering step to determine whether a local feature should merge with a cluster. When the threshold value varies, the derived codebooks and detection results may be significantly different. {The Hough forest approach} \cite{HF09} constructs a codebook (i.e., a tree in a random forest) by using a supervised clustering step that also uses some parameters, such as the depth of a tree and the criteria used to stop the growth of a tree. However, how to choose appropriate values for these parameters remains an open problem. In addition, some popular histogram features, such as Histograms of Oriented Gradient (HOG) \cite{HOG}, are extracted from overlapping image blocks. In that case, the redundancy and the multicollinearity of a derived high-dimensional feature set can be very high, and the performance of object detection approaches can be decreased.

The above issues, i.e., the difficulty of choosing appropriate values for the parameters used in a clustering step and the negative influence of redundancy and multicollinearity, can be solved by using Partial Least Squares (PLS) \cite{Wold1966}. PLS is a popular statistical regression technique, which projects feature vectors onto a much lower dimensional latent subspace. Since the latent components yielded by the projection are mutually orthogonal, the multicollinearity of a feature set is eliminated and the redundancy of the feature set is reduced. As for choosing appropriate values for parameters, the value of the only parameter used in PLS, i.e., the number of latent components, can be determined by using a cross-validation procedure. By exploiting these advantages of PLS, we propose a novel Hough Transform-based object detection approach.  
\begin{algorithm} \label{train}
\caption{The training procedure of the proposed approach}
\KwIn{a set of training images.}
\KwOut{Hough Regression Models $ V_j $.}

\begin{minipage}[t]{0.95\linewidth}
Extract $n$ image patches from the training images and represent the image patches as feature vectors $ \{\mathbf{x}_i\}_{i=1}^n $. Form a matrix $ X_0=(\mathbf{x}_1,\mathbf{x}_2,\cdots,\mathbf{x}_n)^\mathrm{T} $.
\end{minipage}

\begin{minipage}[t]{0.96\linewidth}
Extract $n$ voting vectors $ \{\mathbf{y}_i\}_{i=1}^n $ corresponding to the $n$ feature vectors, respectively. Form a matrix $ Y=(\mathbf{y}_1,\mathbf{y}_2,\cdots,\mathbf{y}_n)^\mathrm{T} $.
\end{minipage}

\begin{minipage}[t]{0.95\linewidth}
For each $ \mathbf{x}_i $, extract a set of $m$ neighbors $\{\mathbf{x}_{i_j}\}_{j=1}^m $ and derive a set of $m+1$ context-encoded feature vectors $ \{\mathbf{x}_{i}-\mathbf{x}_{i_j}\}_{j=0}^m $.
\end{minipage}

\begin{minipage}[t]{0.95\linewidth}
Form $m+1$ context-encoded matrices:\\
\centerline{$ \{ X_j =(\mathbf{x}_{1}-\mathbf{x}_{1_j},\mathbf{x}_{2}-\mathbf{x}_{2_j},\cdots,\mathbf{x}_{n}-\mathbf{x}_{n_j})^\mathrm{T} \}_{j=0}^m $.}
Derive $ m+1 $ context-encoded training sets $ \mathcal{T}=\{ (X_j,Y) \}_{j=0}^m $.
\end{minipage}

\begin{minipage}[t]{0.95\linewidth}
Apply BPLS on each training set to obtain $m+1$ context-encoded Hough Regression Models $ \{ V_j \}_{j=0}^m $.
\end{minipage}

\end{algorithm}
Instead of constructing a codebook as in \cite{ISM08,HF09}, our approach uses Bridge Partial Least Squares (BPLS) \cite{BridgePLS} to establish linear regression models. The obtained models take context-encoded feature vectors as inputs and generate Hough votes for all possible object locations to yield Hough images. We call these linear regression models as context-encoded Hough Regression Models (HRMs). The local maxima of Hough images correspond to the estimated object locations. BPLS is an efficient variant of the traditional PLS technique. The traditional PLS technique uses an inefficient iterative procedure in which an eigenvalue decomposition step is implemented repeatedly to extract enough number of latent components. BPLS can simultaneously extract all latent components for feature vectors by using eigenvalue decomposition only once. The iterative procedure in PLS is not required in BPLS. BPLS was originally proposed in the area of chemometrics and used to analyze functional magnetic resonance imaging (fMRI) data. In this paper, we use BPLS to establish HRMs for object detection. 

Furthermore, we propose a novel multi-scale voting scheme, inspired by the idea of \cite{VotingLines}, to efficiently handle object scale changes. This voting scheme simultaneously casts Hough votes at multiple scales by using only an original image. Therefore, multiple Hough images corresponding to multiple object scales can be obtained simultaneously, while an image pyramid (which is used in \cite{HF09,PSCG,MMHT,MultiHT,LatentHF}) is not required.

\begin{algorithm} \label{test}
\caption{The test procedure of the proposed approach}
\KwIn{a test image and the trained Hough Regression Models.}
\KwOut{object hypotheses at $ S $ different scales $ \{\sigma_s\}_{s=1}^S $.}

\begin{minipage}[t]{0.95\linewidth}
Densely extract all $r$ image patches $ \{\mathrm{p}_\ell\}_{\ell=1}^r $ from the test image.
\end{minipage}

\begin{minipage}[t]{0.95\linewidth}
For each $ \mathrm{p}_\ell $, derive $ m+1 $ context-encoded feature vectors $  \{\mathbf{x}_\ell-\mathbf{x}_{\ell_j}\}_{j=0}^m $.
\end{minipage}

\begin{minipage}[t]{0.95\linewidth}
By using all the Hough Regression Models $ \{ V_j \}_{j=0}^m $ obtained in the training stage, generate $ m+1 $ Hough votes: $ \mathcal{E}_\ell = \{\hat{\mathbf{y}}_{\ell_j}\}_{j=0}^m $ for each $ \mathrm{p}_\ell $. 
\end{minipage}

\begin{minipage}[t]{0.95\linewidth}
At a scale $\sigma_s$, obtain the set of Hough votes cast by all the $r$ image patches: $ \mathcal{A}^{(\sigma_s)}=\bigcup\nolimits_{\ell=1}^{r}\mathcal{E}_\ell^{(\sigma_s)} $.
\end{minipage}

\begin{minipage}[t]{0.95\linewidth}
Obtain $ S $ sets of Hough votes $ \{\mathcal{A}^{(\sigma_s)}\}_{s=1}^S $ and form an $ S $-level Hough image cuboid.
\end{minipage}

\begin{minipage}[t]{0.95\linewidth}
Find all the local maxima at each level of the Hough image cuboid. Accept all the hypotheses corresponding to these local maxima as initial hypotheses.
\end{minipage}

\begin{minipage}[t]{0.95\linewidth}
For any pair of hypotheses $ (h(o,\mathbf{z}_i,\sigma_i),h(o,\mathbf{z}_j,\sigma_j))(i \neq j) $ (where $\sigma_i$ and $\sigma_j$ are the scales of the two hypotheses, respectively), calculate the NPMI between them. If the NPMI is larger than zero, remove the hypothesis whose score is smaller than the other one.
\end{minipage}

\end{algorithm}

Based on this scheme, a principled fusion strategy is proposed to fuse multiple detection hypotheses corresponding to one object to reduce false positives. This strategy reveals and measures the correlation between two hypotheses by evaluating normalized pointwise mutual information (NPMI) \cite{NPMI} between them. If two hypotheses at two different scales are considered to be correlated by evaluating NPMI, they are fused to avoid a false positive.

The proposed approach is called the HRM approach. The training and test procedures of the HRM approach are shown in Algorithm~\ref{train} and \ref{test}, respectively. {In the experiments, the HRM approach is also compared with its several variants to evaluate the influences of its components on its performance.}

This study extends its earlier version, i.e., the PSCG approach \cite{PSCG}, mainly by: (1) exploiting a more efficient variant of PLS, i.e., BPLS, to improve the efficiency in computing HRMs; (2) proposing a novel multi-scale voting scheme to efficiently handle object scale changes and reveal the correlations between hypotheses; (3) generalizing the probabilistic framework in the PSCG approach to describe the proposed multi-scale voting scheme; (4) proposing a principled and NPMI-based strategy to fuse hypotheses to reduce false positives.

The rest of this paper is organized as follows: Section~\ref{relatedwork} summarizes the related work; Section~\ref{HRM} specifies how to establish HRMs by using PLS and BPLS; the proposed multi-scale voting scheme and probabilistic framework are described in Section~\ref{multi-voting}; the NPMI-based fusion strategy is explained in Section~\ref{fusion}; experimental results on popular benchmark datasets are shown in Section~\ref{experiments}; conclusions are given in Section~\ref{conclusion}.

\section{Related Work} \label{relatedwork}

Sliding window and Hough Transform are two major frameworks used in many visual object detection approaches. Owing to the significant work of Leibe et al. \cite{ISM08} (i.e., the ISM approach), the Hough Transform framework becomes more and more popular in detecting irregular-shaped and articulated objects \cite{HF09,4dism,partISM,TUDMotor,FastPRISM,PSCG,MMHT,DGHT,MultiHT,VotingLines,VotingGrouping,LatentHF}. The ISM approach extracts local features from objects in training images. The spacial relationship between each local feature and its corresponding object center is recorded as a voting vector. All the extracted local features are clustered, and the obtained cluster centers form an appearance codebook. Any local feature in a test image is matched to the codebook. The recorded voting vectors corresponding to the activated codebook entries are used to cast Hough votes. The partISM approach \cite{partISM} and voting line-based approach \cite{VotingLines} also build up a codebook by clustering local features. In \cite{partISM}, the parts of a pedestrian are detected independently. The spacial relationship between each part and the center of a pedestrian is learned to predict the locations of pedestrians in test images. 

However, the clustering step for building up a codebook in ISM and partISM is time-consuming. Moreover, in the clustering step, a distance threshold is used to determine whether a local feature should merge with a cluster. The threshold value can significantly influence the effectiveness of the derived codebook. In addition, the weights assigned to all Hough votes cast from a codebook entry are identical in the ISM approach. To address these issues, a number of improvements have been made to the ISM approach in recent years. Instead of unsupervised clustering, the Hough forests \cite{HF09,LatentHF} and DGHT \cite{DGHT} approaches implement a supervised clustering step to construct random forests. The leaf nodes of a tree in a random forest are considered to form a discriminative codebook. Given a local feature in a test image, both its corresponding voting vectors and the probability that the feature belongs to foreground can be obtained from the leaf nodes of a tree. The probability is considered as the weight of any Hough vote cast by the feature. The MMHT \cite{MMHT} approach adapts the probabilistic framework in the ISM approach and learns the weights of Hough votes in a discriminative max-margin framework.

As for handling object scale variations, the ISM \cite{ISM08}, 4D-ISM \cite{4dism}, partISM \cite{partISM}, IRD \cite{TUDMotor}, Fast PRISM \cite{FastPRISM} and MMHT \cite{MMHT} approaches use local feature descriptors to estimate the scales of local features and cast Hough votes in a scale space. The positions where the voting points are most concentrated in the scale space are considered as the locations of object hypotheses. The Hough forests \cite{HF09}, PSCG \cite{PSCG} and latent Hough forest \cite{LatentHF} approaches, as well as the iterative multi-object extraction framework \cite{MultiHT}, simply rescale a test image to form an image pyramid and perform object detection at each level of the pyramid. In the derived Hough image pyramid, 3D local maxima indicate the estimated positions and scales of object hypotheses \cite{HF09}. In \cite{VotingLines}, local scale estimation is considered to be unreliable. Therefore, to solve this problem, voting points are extended to voting lines in \cite{VotingLines}. A voting line consists of the voting points cast by a local feature at all scales in a scale space. The position of an intersection point of voting lines in the scale space indicates the location and scale of an object hypothesis.

{The Partial Least Squares technique \cite{Wold1966} is a supervised dimensional reduction tool, which is usually followed by a feature selection step to discard noisy and redundant features. For instance,  \cite{PLSHuman} and \cite{PLSVehicle} use PLS followed by a feature selection strategy named Ordered Predictors Selection to detect humans and vehicles on challenging datasets, respectively. Recently, a novel pedestrian detection approach \cite{Strengthen} performs feature selection by using an Adaboost algorithm to select spatially pooled covariance matrix features and LBP features. By intensively inspecting the experimental design of the detector, the approach proposed in \cite{Strengthen} obtains an impressive performance in pedestrian detection.} \cite{PLSDataDriven} employs PLS in a multi-stage framework to perform data-driven object detection. Furthermore, a non-linear variant of PLS in Reproducing Kernel Hilbert Space, i.e. Kernel Partial Least Squares \cite{KPLS}, is employed in \cite{PLSHead,PLSPose,PLSAge} to improve the performance of PLS in the tasks of head pose estimation, monocular 3D pose estimation and human age estimation, respectively.

\section{Generating Context-encoded HRMs with BPLS} \label{HRM}

The PLS technique \cite{Wold1966} can reduce redundant information in feature vectors and handle the multicollinearity problem by projecting feature vectors onto a much lower dimensional latent subspace. However, the traditional PLS technique is based on an inefficient iterative procedure which extracts only one latent component in each iteration. BPLS \cite{BridgePLS} is a more efficient variant of PLS, which avoids the iterative procedure and extracts all latent components by using eigenvalue decomposition only once. In this section, we first describe how to utilize PLS to establish context-encoded linear regression models. Each model either predicts object locations or estimates the class labels of image patches.  Afterwards, BPLS is introduced to replace PLS to compute HRMs more efficiently. {Finally, the time complexity of BPLS and PLS is discussed.}

\subsection{Creating Context-Encoded Training Sets}

\newcommand\scct{0.273}
\begin{figure}
  \centering 
  \subfigure[8 adjacent neighboring patches] { \label{Context:a}
  		\includegraphics[scale=\scct]{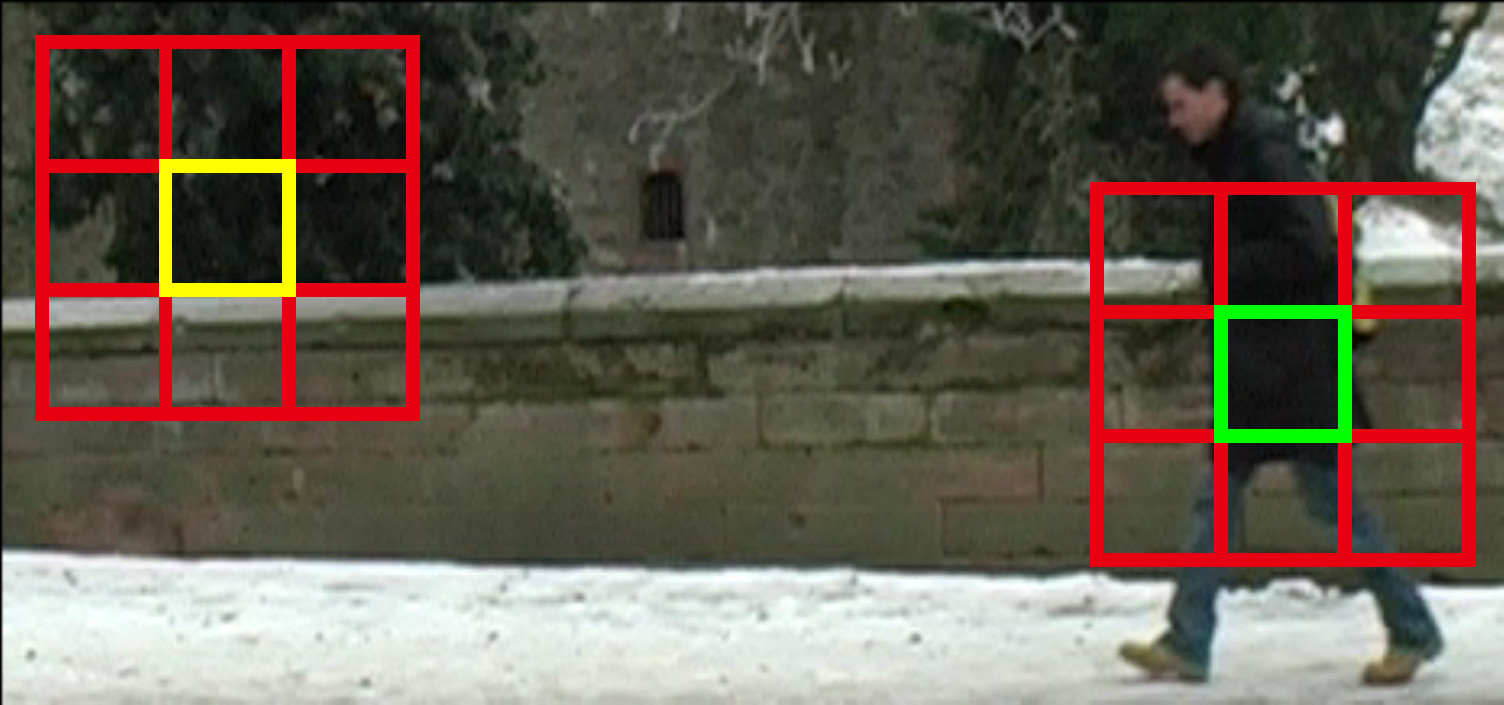}
  	}%
  	\hspace{-0.2mm}
  	\subfigure[8 overlapping neighboring patches] { \label{Context:b}
  	  		\includegraphics[scale=\scct]{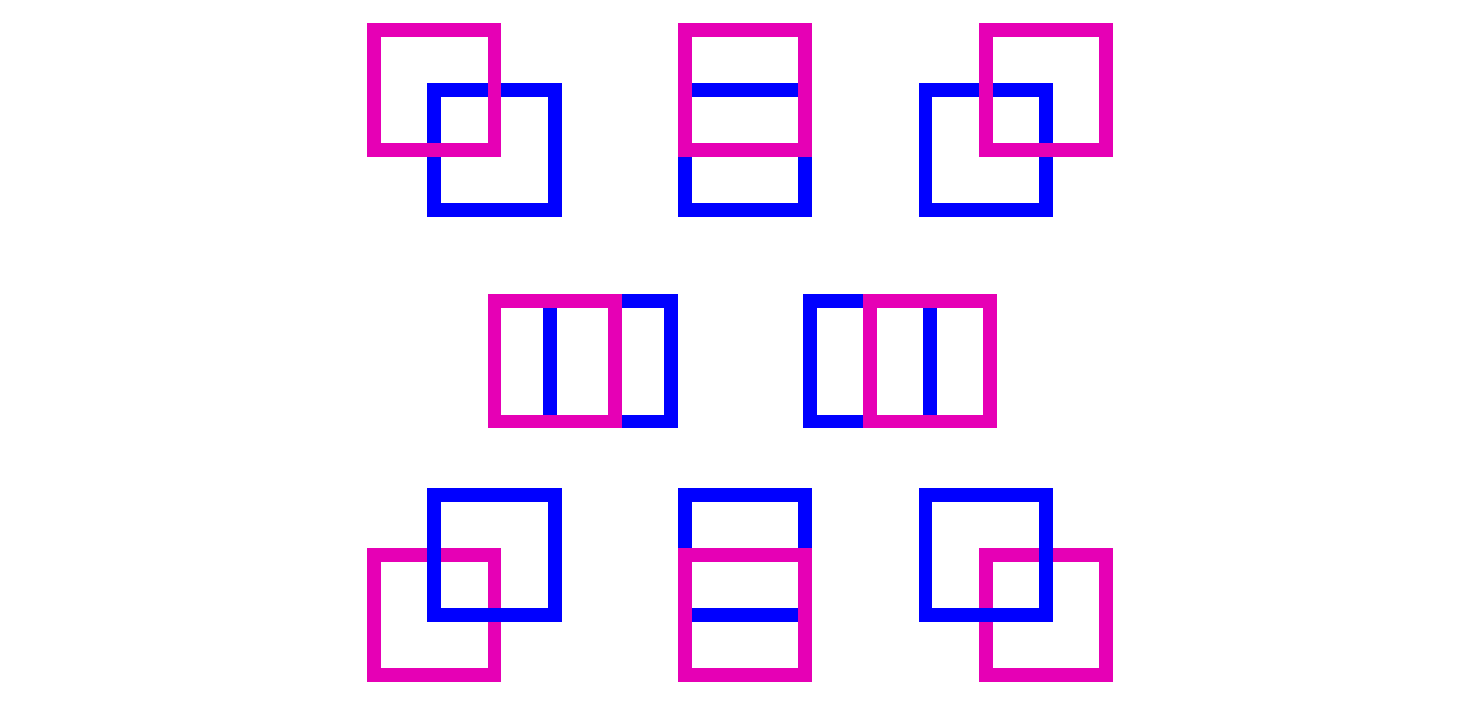}
  	  	}%
  \caption{Demonstration of 16 contextual neighboring patches (8 adjacent neighboring patches (red) of the green and yellow patches (a), and 8 overlapping neighboring patches (pink) of the blue patch (b)). 
  } \label{Context}
\end{figure}%

First, $ n $ small image patches (with a fixed size), denoted as $ \mathcal{P} = \{\mathrm{p}_i\}_{i=1}^n $, are randomly extracted from training images. Half of the image patches are positive samples taken from the bounding boxes corresponding to object locations, and the other ones are negative samples taken from background. The extracted $ n $ image patches are represented as feature vectors $ \{\mathbf{x}_i\}_{i=1}^n $, which form a matrix $ X_0=(\mathbf{x}_1,\mathbf{x}_2,\cdots,\mathbf{x}_n)^\mathrm{T} $. For any positive sample $ \mathrm{p}_i $ in $ \mathcal{P} $, its relative location with respect to its corresponding object center is represented as a two-dimensional vector $ \mathbf{y}_i $. Each $ \mathbf{y}_i $ is called a voting vector. Since the negative samples in $ \mathcal{P} $ are extracted from background, we assign a voting vector $ (-\infty,-\infty)^\mathrm{T} $ to each negative sample. Hence, we obtain $n$ voting vectors $ \{\mathbf{y}_i\}_{i=1}^n $, which form a matrix $ Y=(\mathbf{y}_1,\mathbf{y}_2,\cdots,\mathbf{y}_n)^\mathrm{T} $. Each feature vector $ \mathbf{x}_{i} $ in $X_0$ corresponds to a voting vector $ \mathbf{y}_{i} $ in $Y$.

Context information can help to improve the discriminative ability of an image patch. For example in Fig.~\ref{Context:a}, the green and yellow patches are very similar in appearance. However, the three red patches located on the man's legs can help the green patch to discriminate itself from background and cast reliable Hough votes. In order to extract context information, around each image patch in $ \mathcal{P} $, we further extract a set of  $ m $ neighboring patches (see Fig.~\ref{Context:b} for examples of adjacent and overlapping neighboring patches). The feature vectors of the neighboring patches of an image patch $ \mathrm{p}_i $ in $ \mathcal{P} $ are denoted as $ \mathcal{N}_i=\{\mathbf{x}_{i_j}\}_{j=1}^m $. By repeatedly subtracting $ \mathbf{x}_{i_j} $ from $ \mathbf{x}_i $ (as $ j $ varies from $ 1 $ to $ m $), a set of $ m+1 $ \textit{context-encoded feature vectors} can be derived for the image patch $ \mathrm{p}_i $: 
\begin{equation}
	\mathcal{D}_i=\{\mathbf{x}_{i}-\mathbf{x}_{i_j}\}_{j=1}^m \bigcup \{\mathbf{x}_i\}=\{\mathbf{x}_{i}-\mathbf{x}_{i_j}\}_{j=0}^m
\end{equation}
where $\mathbf{x}_{i_0}=\mathbf{0}$.

Now for all the image patches $ \{ \mathrm{p}_i \}_{i=1}^n $ in $ \mathcal{P} $, we have $ n $ sets $ \{ \mathcal{D}_i \}_{i=1}^n $, and each $ \mathcal{D}_i $ contains $ m+1 $ context-encoded feature vectors. As a result, $ m+1 $ matrices can be formed as follows:
\begin{equation}
	\{ X_j =(\mathbf{x}_{1}-\mathbf{x}_{1_j},\mathbf{x}_{2}-\mathbf{x}_{2_j},\cdots,\mathbf{x}_{n}-\mathbf{x}_{n_j})^\mathrm{T} \}_{j=0}^m.
\end{equation}
Note that for each matrix $ X_j $, the $ i $-th row is the transpose of the $ j $-th vector in $ \mathcal{D}_i $. Recall that each feature vector $ \mathbf{x}_{i} $ in $X_0$ corresponds to a voting vector $ \mathbf{y}_{i} $ in $Y$. We let each $ \mathbf{x}_{i} $ share its corresponding $ \mathbf{y}_i $ with all the context-encoded feature vectors in $\mathcal{D}_i$. Thus, $ m+1 $ context-encoded training sets are obtained as $ \mathcal{T}=\{ (X_j,Y) \}_{j=0}^m $.

\subsection{Generating HRMs with Partial Least Squares} \label{traPLS}

With each training set $ (X_j,Y) $ in $ \mathcal{T} $, we use PLS to establish a linear regression model $ V_j $ as follows:
\begin{equation}
V_j: \tilde{Y}=\tilde{X}_jB_j+R_j,
\end{equation}
where $ \tilde{X}_j $ and $ \tilde{Y} $ are the mean-centered variants of $ X_j $ and $ Y $, respectively; $B_j$ is a matrix of regression coefficients and $R_j$ is a residual matrix. From \cite{FormulaOf_B,FormulaOf_B'}, the matrix of regression coefficients $ B_j $ can be estimated as follows:
\begin{equation}
	B_j=W_j(T_j^\mathrm{T}\tilde{X}_jW_j)^{-1}T_j^\mathrm{T}\tilde{Y},
\end{equation}
where $ W_j=(w_1,w_2,...,w_c) $ is a matrix of weights, and $ \{w_k\}_{k=1}^c $ in $ W_j $ are the weight vectors in PLS; $ T_j=(t_1,t_2,...,t_c) $ is a matrix of scores. $ \{t_k\}_{k=1}^c $ in $ T_j $ are the score vectors, and are also the latent components for the feature vectors in $ \tilde{X}_j $. These latent components span a $c$-dimensional latent space. The dimensionality of the latent space (i.e., $c$) is much lower than the dimensionality of the feature vectors in $ \tilde{X}_j $. $ c $ is the only parameter used in PLS and can be estimated by using a cross-validation procedure.

To obtain each weight vector $ w_k $ and score vector $ t_k $, traditional variants of PLS usually conduct an iterative procedure as described in Algorithm~\ref{PLSalg} \cite{Wold1966,FormulaOf_B,FormulaOf_B'}. Step 3 in Algorithm~\ref{PLSalg} indicates that $w_k$ is the first dominant eigenvector of the matrix $ E_k^\mathrm{T}F_k^{}F_k^\mathrm{T}E_k^{} $. Therefore, in each iteration in Algorithm~\ref{PLSalg}, an eigenvalue decomposition step (step 3) followed by the deflations of the matrices $ E_k $ and $ F_k $ (step 5) is computed. Once all the weight vectors $ \{w_k\}_{k=1}^c $ and the score vectors $ \{t_k\}_{k=1}^c $ are obtained, we can have the linear regression model $ V_j $ as described in step 7 - step 9 of Algorithm~\ref{PLSalg}.

\begin{algorithm} \label{PLSalg}
\caption{A typical iterative procedure used in traditional Partial Least Squares algorithms}
\KwIn{a predictor matrix $ \tilde{X}_j $ and a response matrix $ \tilde{Y} $.}
\KwOut{\justifying a matrix of weights $ W_j=(w_1,w_2,...,w_c)$, a matrix of scores $T_j=(t_1,t_2,...,t_c) $, a matrix of regression coefficients $ B_j $, a residual matrix $ R_j $, and a Hough Regression Model $ V_j $.}

Initialize:\\
\centerline{$ E_1=\tilde{X}_j $, $ F_1=\tilde{Y} $.}

\For {$k=1$ to $c$}
{
Calculate the first dominant eigenvector of $E_k^\mathrm{T}F_k^{}F_k^\mathrm{T}E_k$ to obtain the $k$-th weight vector $w_k$:\\
\centerline{$w_k^{}\,=\,\underset{w}{\mathrm{argmax}} \ \ w^\mathrm{T}E_k^\mathrm{T}F_k^{}F_k^\mathrm{T}E_k^{}w, \ \ s.t.:  w_{}^\mathrm{T}w=1.$}

Calculate the $k$-th score vector:\\
\centerline{$t_k = E_k * w_k$.}

Deflate the matrices $ E_k^{} $ and $ F_k^{} $:\\
\centerline{$ E_{k+1}^{} = E_k^{} - t_k^{}t_k^\mathrm{T}E_k^{} $, \, $ F_{k+1}^{} = F_k^{} - t_k^{}t_k^\mathrm{T}F_k^{} $.}
}

Calculate the matrix of regression coefficients:\\
\centerline{$ B_j = W_j(T_j^\mathrm{T}\tilde{X}_jW_j)^{-1}T_j^\mathrm{T}Y $.}

Calculate the residual matrix:\\
\centerline{$ R_j = \tilde{Y}-\tilde{X}_jB_j $.}

Obtain the Hough Regression Model $ V_j $:\\
\centerline{$ V_j: \tilde{Y}=\tilde{X}_jB_j+R_j $.}
\end{algorithm} 

Algorithm~\ref{PLSalg} specifies how to obtain the model $ V_j $ by using the training set $ (X_j,Y) $ in $ \mathcal{T} $. Therefore, when all the $ m+1 $ training sets in $ \mathcal{T} $ are used respectively, we can obtain $ m+1 $ models: $ \mathcal{V}=\{ V_j \}_{j=0}^m $. All the models are used to generate Hough votes to predict object locations. Let us consider a test image containing $r$ image patches $ \{\mathrm{p}^{}_\ell\}_{\ell=1}^{r} $. For each $ \mathrm{p}^{}_\ell $, we derive $ m+1 $ context-encoded feature vectors: $  \mathcal{D}_\ell=\{\mathbf{x}_\ell-\mathbf{x}_{\ell_j}\}_{j=0}^m $. The $j$-th vector $ ( \mathbf{x}_\ell-\mathbf{x}_{\ell_j} ) $ corresponds to the $j$-th model $ V_j $ in $ \mathcal{V} $. By using the vector $ ( \mathbf{x}_\ell-\mathbf{x}_{\ell_j} ) $ and the matrix of regression coefficients $B_j$ in the model $V_j$,  the image patch $ \mathrm{p}^{}_\ell $ produces a voting vector as follows:
\begin{equation}
	\mathbf{\hat{y}}_{\ell_j}=\dfrac{1}{n}\sum_{i=1}^{n}\mathbf{y}_i + B_j^\mathrm{T}\Big[( \mathbf{x}_\ell - \mathbf{x}_{\ell_j} )-\dfrac{1}{n}\sum_{i=1}^{n}(\mathbf{x}_i-\mathbf{x}_{i_j})\Big]. \label{estimation}
\end{equation}
Thus, by varying $j$ from $0$ to $m$, each image patch $ \mathrm{p}^{}_\ell $ can produce $ m+1 $ voting vectors (i.e., Hough votes): $ \mathcal{E}_\ell = \{\hat{\mathbf{y}}_{\ell_j}\}_{j=0}^m $. Each voting vector $ \mathbf{\hat{y}}_{\ell_j} $ indicates a possible object location relative to the image patch $ \mathrm{p}^{}_\ell $. Since the $ m+1 $ linear regression models $ \{ V_j \}_{j=0}^m $ in $ \mathcal{V} $ are established by using the context-encoded training sets in $ \mathcal{T} $ and are used to produce Hough votes, we call these models as \textit{context-encoded Hough Regression Models (HRMs)}. 

In addition, we also establish $m+1$ \textit{context-encoded Label Regression Models (LRMs)}: $ \mathcal{L}=\{L_j\}_{j=0}^m $. To obtain the LRMs, for each training image patch $ \mathrm{p}_i $ in $ \mathcal{P} $, its voting vector $ \mathbf{y}_i $ in $ Y $ is replaced with its class label $ y_i \in \{+1,-1\} $. Then, the LRMs are computed by repeating the regression process described in Algorithm~\ref{PLSalg}. Similar to Eq.~(\ref{estimation}), by using all the $m+1$ LRMs in $ \mathcal{L} $, we can obtain $ m+1 $ estimated class labels for the test image patch $ \mathrm{p}^{}_\ell $: $ \mathcal{C}_\ell = \{\hat{y}_{\ell_j}\}_{j=0}^m$.

\subsection{Generating HRMs with Bridge Partial Least Squares}

As can be seen in each iteration in Algorithm~\ref{PLSalg}, only one weight vector $ w_k $ is extracted by using an eigenvalue decomposition step (step 3). Afterwards, a matrix deflation step (step 5) is also employed in each iteration. These iterative steps seriously decrease the efficiency of PLS. The matrix deflation step after extracting each weight vector is necessary because the matrix $ E_1^\mathrm{T}F_1^{}F_1^\mathrm{T}E_1^{} $ (i.e., $ \tilde{X}_j^\mathrm{T}\tilde{Y}\tilde{Y}^\mathrm{T}\tilde{X}_j $) is rank-deficient. The rank of $\tilde{X}_j^\mathrm{T}\tilde{Y}\tilde{Y}^\mathrm{T}\tilde{X}_j$ is not larger than the rank of $ \tilde{Y} $. In our case, we have $rank(\tilde{X}_j^\mathrm{T}\tilde{Y}\tilde{Y}^\mathrm{T}\tilde{X}_j) \le rank(\tilde{Y}) = 2$, since the voting vectors in $ \tilde{Y} $ are two-dimensional. Assume that we want to extract $c$ latent components $ \{t_k\}_{k=1}^c $ as in Algorithm~\ref{PLSalg}. If the first $c$ dominant eigenvectors of $ \tilde{X}_j^\mathrm{T}\tilde{Y}\tilde{Y}^\mathrm{T}\tilde{X}_j $ are extracted simultaneously, the $3$rd to the $c$-th dominant eigenvectors can not explain the covariance between the matrices $ \tilde{X}_j $ and $ \tilde{Y} $. Consequently, only two latent components (i.e., $ t_1 $ and $ t_2 $ in Algorithm~\ref{PLSalg}) can be extracted \cite{BridgePLS,EfficientPLS}. Thus, in order to obtain more latent components, the traditional PLS technique solves the rank deficiency problem by employing the matrix deflation step (step 5 in Algorithm~\ref{PLSalg}). As a result, the eigenvalue decomposition step (step 3) and the matrix deflation step (step 5) in Algorithm~\ref{PLSalg} have to be repeated in each iteration, which is time-consuming.

BPLS \cite{BridgePLS} can solve the rank deficiency problem and extract all required latent components simultaneously in one eigenvalue decomposition step. Thus, BPLS is more efficient than PLS. The main idea of BPLS is to introduce a ridge-parameter $\alpha$ into $ \tilde{X}_j^\mathrm{T}\tilde{Y}\tilde{Y}^\mathrm{T}\tilde{X}_j $ as follows:
\begin{align}
   M &=  \tilde{X}_j^\mathrm{T}(\alpha I + (1-\alpha)\tilde{Y}\tilde{Y}^\mathrm{T})\tilde{X}_j \\
	 &=  \alpha \tilde{X}_j^\mathrm{T}\tilde{X}_j + (1-\alpha) \tilde{X}_j^\mathrm{T}\tilde{Y}\tilde{Y}^\mathrm{T}\tilde{X}_j \label{tradeoff} \\
	 &=  \begin{bmatrix}
	  	  \sqrt{\alpha}\,\tilde{X}_j^\mathrm{T} & \sqrt{1-\alpha}\,\tilde{X}_j^\mathrm{T}\tilde{Y}
	 	  \end{bmatrix}
	 	  \begin{bmatrix}
	 	  	  \sqrt{\alpha}\,\tilde{X}_j^\mathrm{T} \\[2mm] \sqrt{1-\alpha}\,\tilde{X}_j^\mathrm{T}\tilde{Y}
	 	  \end{bmatrix},
\end{align}
where $\alpha \in [0,1] $ and $I$ is an identity matrix. When $\alpha$ is a very small number, the matrix $ M $ highly approximates the matrix $ \tilde{X}_j^\mathrm{T}\tilde{Y}\tilde{Y}^\mathrm{T}\tilde{X}_j $. Therefore, replacing $ \tilde{X}_j^\mathrm{T}\tilde{Y}\tilde{Y}^\mathrm{T}\tilde{X}_j $ with $ M $ in PLS does not affect the effectiveness of the established model $ V_j $ \cite{EfficientPLS}. This replacement leads to a PLS regression when $\alpha=0$, and yields a principal components regression (PCR) when $\alpha=1$ \cite{BridgePLS,EfficientPLS}.

By linear algebra, we have:
\begin{algorithm} \label{BridgePLS}
\caption{The Bridge Partial Least Squares algorithm}
\KwIn{a predictor matrix $ \tilde{X}_j $ and a response matrix $ \tilde{Y} $.}
\KwOut{\justifying a matrix of weights $ W_j=(w_1,w_2,...,w_c)$, a matrix of scores $T_j$, a matrix of regression coefficients $ B_j $, a residual matrix $ R_j $, and a Hough Regression Model $ V_j $.}

Introduce the ridge-parameter $\alpha$ into $ \tilde{X}_j^\mathrm{T}\tilde{Y}\tilde{Y}^\mathrm{T}\tilde{X}_j $:\\
\centerline{$ M = \tilde{X}_j^\mathrm{T}(\alpha I + (1-\alpha)\tilde{Y}\tilde{Y}^\mathrm{T})\tilde{X}_j $.}

Calculate the first $c$ dominant eigenvectors of $M$ to obtain the matrix of weights $W_j$:\\
\centerline{$ W_j = \underset{W}{\mathrm{argmax}} \ \ \mathrm{Tr}(W^\mathrm{T}MW), \ \ s.t.:  W^\mathrm{T}W=1 $.}

Calculate the matrix of scores:\\
\centerline{$ T_j = \tilde{X}_j W_j $.}

Calculate the matrix of regression coefficients:\\
\centerline{$ B_j = W_j(T_j^\mathrm{T}T_j)^{-1}T_j^\mathrm{T}Y $.}

Calculate the residual matrix:\\
\centerline{$ R_j = \tilde{Y}-\tilde{X}_jB_j $.}

Obtain the Hough Regression Model $ V_j$:\\
\centerline{$ V_j: \tilde{Y}=\tilde{X}_jB_j+R_j $.}
\end{algorithm}
\begin{equation}
	\begin{split}
		rank(M) &= rank\left(
					  	  \begin{bmatrix}
						  	  \sqrt{\alpha}\,\tilde{X}_j^\mathrm{T} &  \sqrt{1-\alpha}\,\tilde{X}_j^\mathrm{T}\tilde{Y}
					  	  \end{bmatrix}
				  	  \right)\\
				&= rank(\tilde{X}_j).
	\end{split}
\end{equation}
Hence, different from $ \tilde{X}_j^\mathrm{T}\tilde{Y}\tilde{Y}^\mathrm{T}\tilde{X}_j $, the key property of $M$ is that $rank(M)$ is equal to $rank(\tilde{X}_j)$. Recall that we assume that $c$ latent components $ \{t_k\}_{k=1}^c $ are to be extracted. In practice, $rank(M)$ (i.e., $rank(\tilde{X}_j)$) is usually comparatively large, and it is usually much larger than $c$. Therefore, all the first $c$ dominant eigenvectors of $M$ are able to explain the covariance of $M$. By replacing $ \tilde{X}_j^\mathrm{T}\tilde{Y}\tilde{Y}^\mathrm{T}\tilde{X}_j $ with $ M $, the steps of deflating matrix (step 5) and repeatedly computing eigenvalue decomposition (step 3) in Algorithm~\ref{PLSalg} are no longer necessary. Algorithm~\ref{BridgePLS} \cite{BridgePLS,EfficientPLS} shows the procedure of BPLS. From Algorithm~\ref{BridgePLS} we can see that BPLS simultaneously extracts the first $c$ dominant eigenvectors of $M$ by using eigenvalue decomposition only once instead of the iterative procedure used in PLS. The $c$ eigenvectors are the obtained weight vectors $ \{w_k\}_{k=1}^c$ (see step 2 in Algorithm~\ref{BridgePLS}).

As with Subsection~\ref{traPLS}, when all the $m+1$ training sets in $ \mathcal{T} $ are used respectively, we can obtain $ m+1 $ HRMs and $ m+1 $ LRMs by using BPLS. Compared to the traditional PLS technique, BPLS uses eigenvalue decomposition only once and can avoid the matrix deflation step. Therefore, the efficiency of BPLS is higher than that of PLS. 

\subsection{{Discussions on Time Complexity}}
{
As can be seen from Algorithms~{\ref{PLSalg}} and {\ref{BridgePLS}}, the computational burden in PLS and BPLS mainly concentrates on the eigenvalue decomposition steps. Assuming that the predictor matrix $ \tilde{X}_j $ contains $ \hat{n} $ columns, the time complexity of the eigenvalue decomposition of $ E_k^\mathrm{T}F_k^{}F_k^\mathrm{T}E_k $ (i.e., an $\hat{n}\times\hat{n}$ matrix) in step 3 of Algorithm~{\ref{PLSalg}} is $O(\hat{n}^3)$. Since step 3 is executed $c$ times in Algorithm~{\ref{PLSalg}}, the time complexity of the traditional PLS technique is $\sim O(c*\hat{n}^3)$ ($c$ is the number of latent components and is determined by using a cross validation procedure). As for BPLS, the eigenvalue decomposition of $M$ (i.e., an $\hat{n}\times\hat{n}$ matrix) in step 2 of Algorithm~{\ref{BridgePLS}} also has a time complexity of $O(\hat{n}^3) $. However, since step 2 is executed only once in Algorithm~{\ref{BridgePLS}}, the time complexity of the BPLS algorithm is only $\sim O(\hat{n}^3)$. Thus, the time complexity can be reduced from $O(c*\hat{n}^3)$ to $O(\hat{n}^3)$ by using BPLS instead of PLS. When $\hat{n}$ is a large number, the efficiency of the training stage of the proposed HRM approach can be significantly improved.}

\section{Multi-scale Voting Scheme} \label{multi-voting}

In this section, we first discuss the characteristics of two existing solutions to the problem of object scale variations: image pyramids and voting lines. By inheriting the advantages and overcoming the disadvantages of these two solutions, we propose a novel multi-scale voting scheme that simultaneously casts Hough votes at multiple scales by only using an original image. Thus, image pyramids, which are widely used in modern detection algorithms \cite{HOG,HF09,PSCG,MMHT,MultiHT,LatentHF}, are not required. It is of more importance that the proposed voting scheme serves as a foundation for revealing the correlations between hypotheses at different scales. This foundation leads to a principled and NPMI-based fusion strategy proposed in Section~\ref{fusion} which can reduce false positives.

\subsection{Image Pyramids} \label{imagepyra}

\begin{figure}%
 	\centering%
 	\subfigure[Image pyramid]{ \label{fp:a}
 	\begin{overpic}[width=0.475\linewidth]{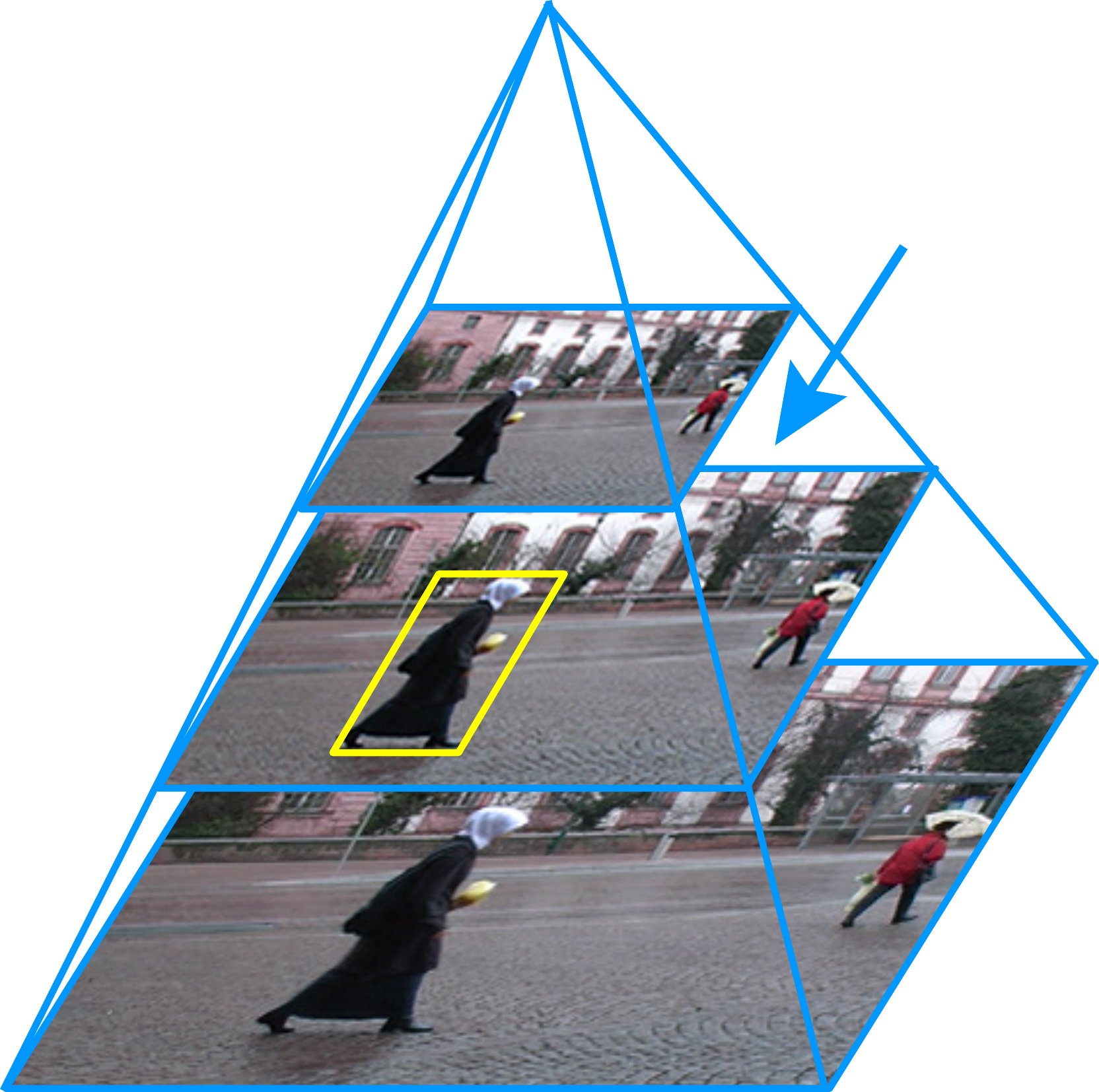}
 	            \put(\LenToUnit{0.35\linewidth},\LenToUnit{0.38\linewidth}){\footnotesize{Original image}}
 	          \end{overpic}
 	}
 	\subfigure[Hough image pyramid]{ \label{fp:b}
 	\includegraphics[width=0.475\linewidth]{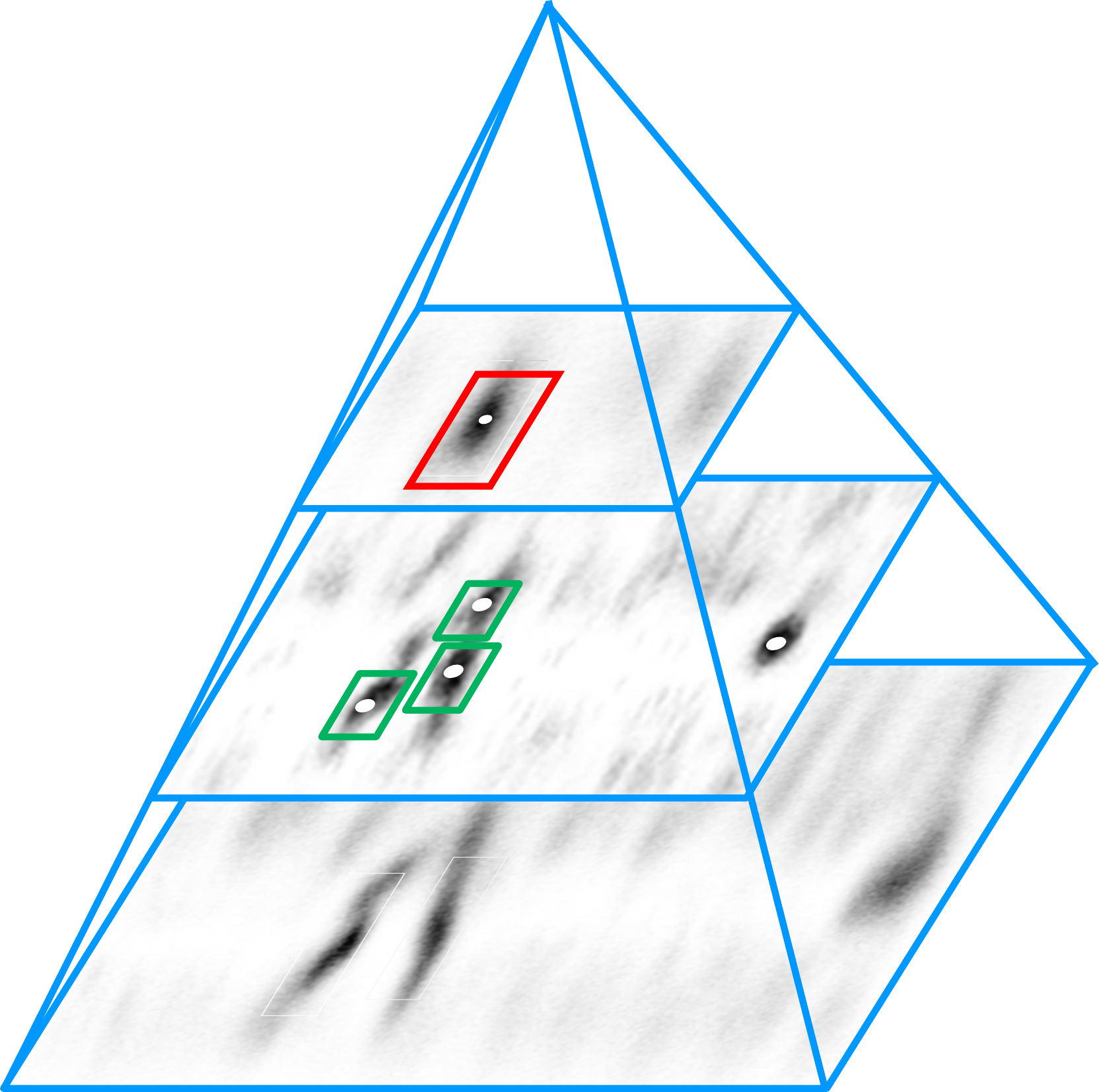}
 	}
 \caption{Illustration of an image pyramid (a) and the corresponding Hough image pyramid (b). In (b), the gray points represent the voting points cast by the image patches extracted from (a); the red and green boxes indicate the true and false positive hypotheses corresponding to the object in the yellow box in (a), respectively.}
 \label{fp}
 \end{figure}

As we said before, image pyramids are widely used to handle object scale changes \cite{HOG,HF09,PSCG,MMHT,MultiHT,LatentHF}. For example in Fig.~\ref{fp:a}, a 3-level image pyramid is formed by resizing the original image. By applying any Hough Transform-based detector at the three levels of the image pyramid, the image features extracted from the three levels cast votes to form three Hough images, respectively, as shown in Fig.~\ref{fp:b}. The voting points are represented as gray points, and the Hough images form a Hough image pyramid. Two problems exist in using an image pyramid to handle object scale changes:
\begin{itemize}
	\item Object detection is performed repeatedly at each level of an image pyramid, which is greatly time-consuming.
	\item Object hypotheses obtained at different levels of a Hough image pyramid are derived from different levels of an image pyramid. This may lead to a result that multiple true and false positive hypotheses corresponding to one object are all accepted as different detected objects.
\end{itemize}

 \newcommand\wa{0.475\linewidth}
 
The first problem is apparent. As for the second one, let us consider the object in the yellow box in Fig.~\ref{fp:a}. By seeking the positions where the voting points are most clustered at the three levels of the Hough image pyramid in Fig.~\ref{fp:b}, it can be seen that the object in the yellow box corresponds to four hypotheses in the Hough image pyramid. The three hypotheses in the green boxes are false positives, and the hypothesis in the red box is a true positive. The false positives may be caused by intra-class variations, background noise, etc. The Hough votes for the false positives are cast by the image features at the second level of the image pyramid in Fig.~\ref{fp:a}, while the votes for the true positive are cast by the image features at the first level of the image pyramid. In other words, although the true positive and the three false positive hypotheses in Fig.~\ref{fp:b} correspond to the same object, they are derived from different levels of the image pyramid. Therefore, it is difficult to reveal or measure the correlations between the true positive and the three false positives. They may be accepted as different detected objects. Some approaches (e.g., \cite{ISM08,VotingLines,MMHT}) use time-consuming verification steps to identify false positives. The Hough forest approach \cite{HF09} proposes to seek the 3D local maxima in a Hough image pyramid to fuse multiple hypotheses corresponding to one object. However, this strategy is still ineffective for the four hypotheses in Fig.~\ref{fp:b}, because each of them can be considered as a 3D local maximum in the Hough image pyramid.

\subsection{Voting Lines}

\begin{figure}%
	\centering%
	\subfigure[A voting line in a scale space]{ \label{vl:a}
	\begin{overpic}[width=0.5\linewidth]{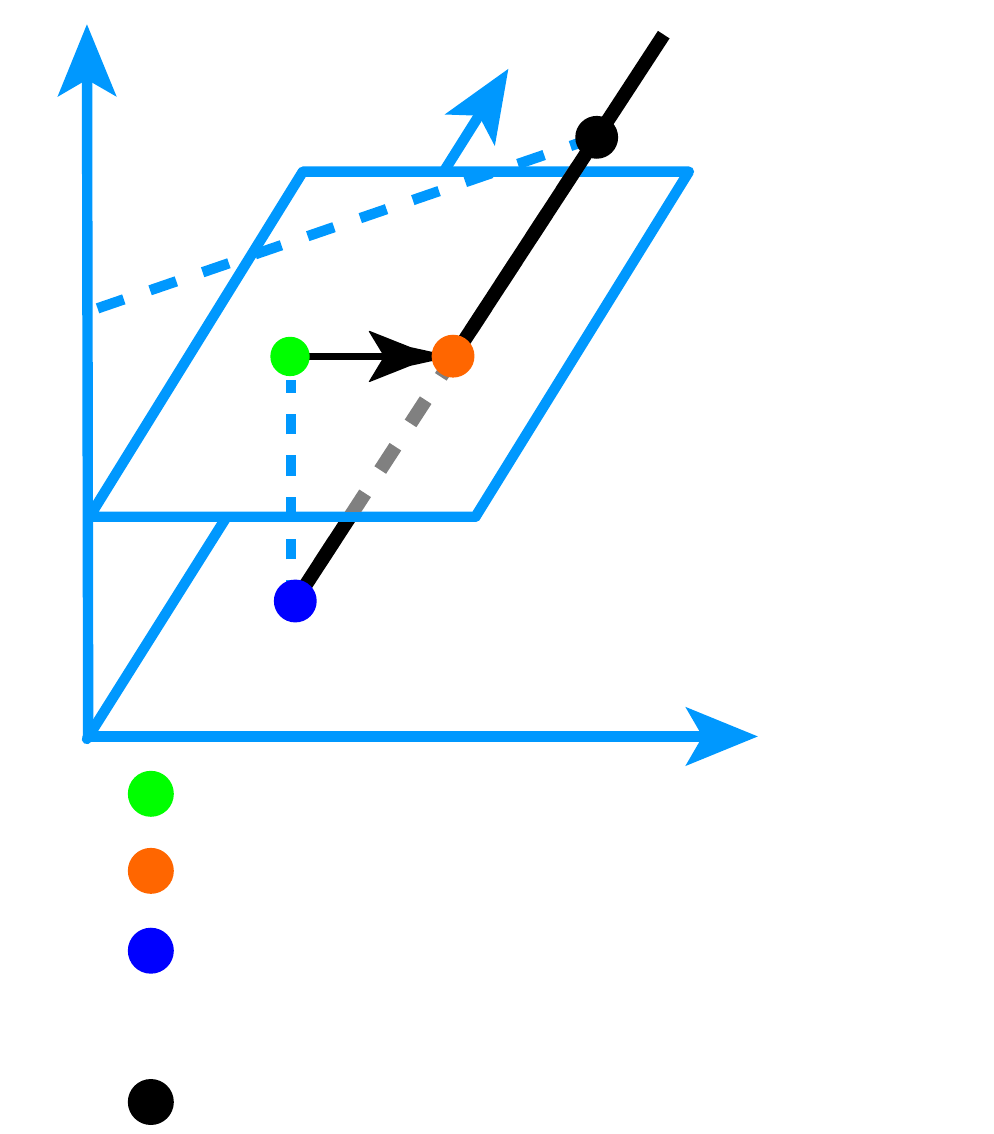}
		\put(\LenToUnit{0.24\linewidth},\LenToUnit{0.54\linewidth}){\small{$\bm{y}$}}
		\put(\LenToUnit{0.38857\linewidth},\LenToUnit{0.19\linewidth}){\small{$\bm{x}$}}
		\put(0,\LenToUnit{0.5257\linewidth}){\small{$\bm\sigma$}}
		\put(0,\LenToUnit{0.4\linewidth}){\small{${\sigma_s}$}}
		\put(0,\LenToUnit{0.3\linewidth}){\small{${\sigma_0}$}}
		\put(\LenToUnit{0.01\linewidth},\LenToUnit{0.18857\linewidth}){\small{$\bm{O}$}}
		\put(\LenToUnit{0.065\linewidth},\LenToUnit{0.317\linewidth}){\small{$\Omega_0$}}
		\put(\LenToUnit{0.062857\linewidth},\LenToUnit{0.205\linewidth}){\small{$\Omega$}}
		\put(\LenToUnit{0.097\linewidth},\LenToUnit{0.36\linewidth}){\footnotesize{${(x,y,\sigma_0)}$}}
		\put(\LenToUnit{0.2457\linewidth},\LenToUnit{0.38\linewidth}){\footnotesize{${(x+d,y,\sigma_0)}$}}
		\put(\LenToUnit{0.315\linewidth},\LenToUnit{0.494\linewidth}){\footnotesize{${(x+\sigma_s/\sigma_0*d,y,\sigma_s)}$}}
		\put(\LenToUnit{0.3143\linewidth},\LenToUnit{0.56\linewidth}){\footnotesize{Voting line}}
		\put(\LenToUnit{0.1143\linewidth},\LenToUnit{0.24\linewidth}){\footnotesize{$(x,y,0)$}}
		\put(\LenToUnit{0.257\linewidth},\LenToUnit{0.297\linewidth}){\footnotesize{Image plane}}
		\put(\LenToUnit{0.09143\linewidth},\LenToUnit{0.16\linewidth}){\footnotesize{: An image feature}}
		\put(\LenToUnit{0.09143\linewidth},\LenToUnit{0.12143\linewidth}){\footnotesize{: A voting point}}
		\put(\LenToUnit{0.09143\linewidth},\LenToUnit{0.062857\linewidth}){\footnotesize{\makecell[l]{: The projection of the green\\~~\hspace{1pt}point on the plane $\Omega$}}}
		\put(\LenToUnit{0.09143\linewidth},\LenToUnit{0.0057\linewidth}){\footnotesize{: A point on the voting line}}
	\end{overpic}
	}
	\subfigure[A cluster of voting lines]{ \label{vl:b}
	\begin{overpic}[width=0.45\linewidth]{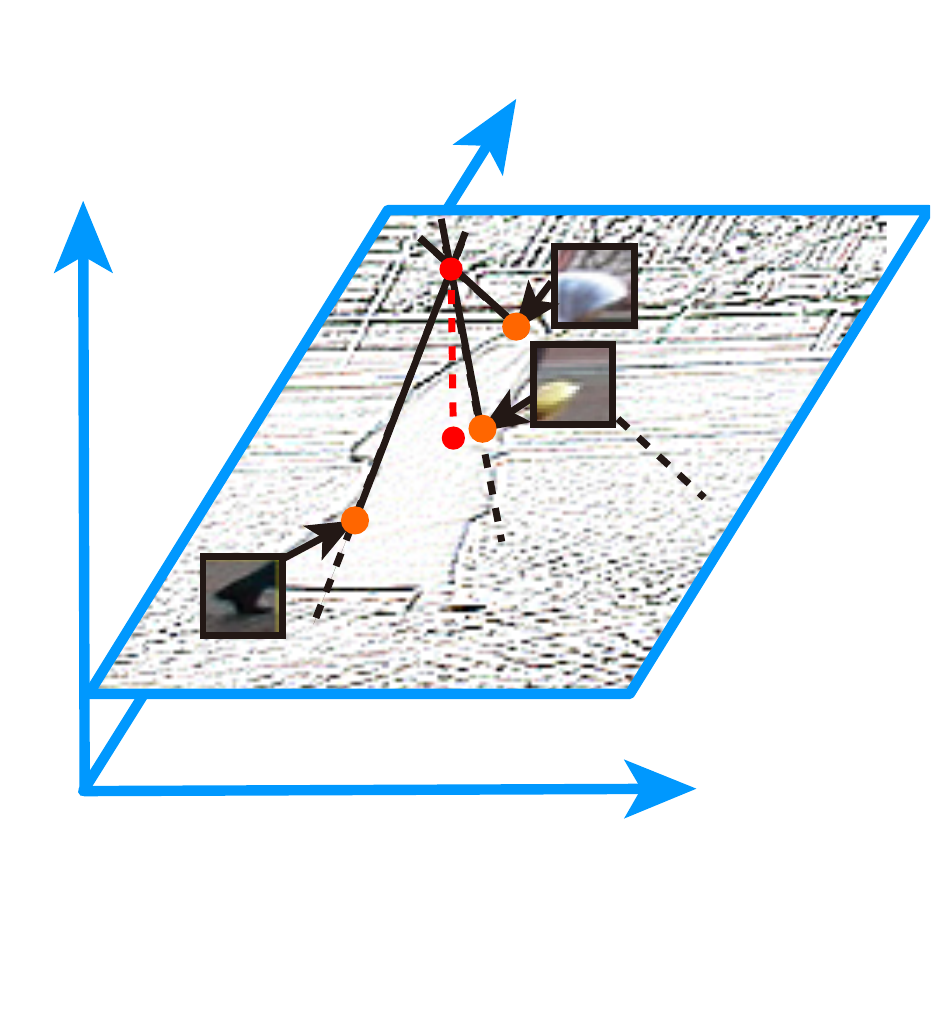}
		\put(0,\LenToUnit{0.37\linewidth}){\small{$\bm\sigma$}}
		\put(\LenToUnit{-0.00493\linewidth},\LenToUnit{0.15787\linewidth}){\small{${\sigma_0}$}}
		\put(\LenToUnit{0.32\linewidth},\LenToUnit{0.12827\linewidth}){\small{$\bm{x}$}}
		\put(\LenToUnit{0.23\linewidth},\LenToUnit{0.45387\linewidth}){\small{$\bm{y}$}}
	\end{overpic}
	}
\caption{Illustration of the basic idea of voting lines. In (a), an image feature casts a voting line. In (b), we use the original image in Fig.~\ref{fp:a} but we only show the contours for better viewing. Three local image features (enclosed by three black boxes) are extracted from an object. Each image feature casts a voting line. Ideally, the three voting lines intersect at a single point. The projection of the intersection point on the image plane indicates the center of the object. See text for more details.}
\label{vl}
\end{figure}
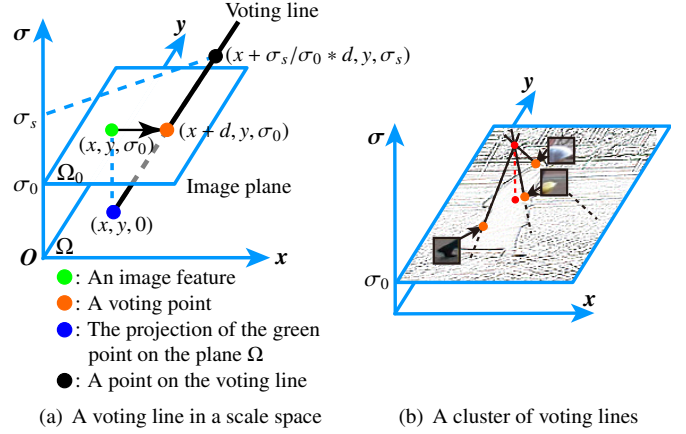

Bj\"{o}rn Ommer et al. \cite{VotingLines} indicate that local scales estimated by local feature descriptors are unreliable. In order to obtain the locations and scales of objects without using unreliable local scale estimates, voting points are generalized to voting lines in a scale space \cite{VotingLines}. The basic idea of \cite{VotingLines} is depicted in Fig.~\ref{vl}. In the training stage of most Hough Transform-based approaches (e.g. \cite{ISM08,MultiHT,HF09}), the training images are resized beforehand so that the scales of the objects used for training are identical. Let us denote the scales of the objects used in the training stage as $ \sigma_0 $. A test image is placed on the plane $ \Omega_0$, whose equation is $ \sigma=\sigma_0 $, in the scale space in Fig.\ref{vl:a} for better understanding. During testing, an image feature, which is represented as the green point in Fig.~\ref{vl:a}, is extracted from the test image at the coordinates $ (x,y,\sigma_0) $ in the scale space. In most Hough Transform-based approaches, this feature may cast a 2D Hough vote along the plane $ \Omega_0$ to the orange point at the coordinates $ (x+d,y,\sigma_0) $ in Fig.~\ref{vl:a}. However, the 2D Hough vote is generalized to a 3D Hough vote, i.e., a voting line, in \cite{VotingLines}. The voting line produced by the image feature at the coordinates $ (x,y,\sigma_0) $ (the green point) in Fig.~\ref{vl:a} is determined by two points in the scale space: the first one is the orange voting point at the coordinates $ (x+d,y,\sigma_0) $; the second one is the blue point at the coordinates $ (x,y,0) $, which is the projection of the green point on the plane $ \Omega $ (i.e., $\sigma = 0$). A voting line can be regarded as an infinite number of voting points, where each point corresponds to a unique scale in the scale space. For example, the black point on the voting line in Fig.~\ref{vl:a} corresponds to a scale $ \sigma_s $. According to geometry, we can simply compute the coordinates of the black point, which is equal to $ (x+\sigma_s/\sigma_0*d,y,\sigma_s) $.

Ideally, the voting lines produced by the image features on one object would intersect at a single point, as shown in Fig.~\ref{vl:b}. The coordinates of the intersection point indicate the position and scale of a hypothesis. However, in practice, due to the factors such as intra-class variations, background noise, etc., the intersection point usually degrades to a scattered point cloud instead of an ideal single point. Therefore, a clustering algorithm should be employed to cluster the voting lines. The scattered cloud of each cluster corresponds to a hypothesis. 

This voting line-based approach can avoid detecting objects repeatedly at several scales, which is required by an image pyramid-based approach. Moreover, since all points on a voting line are considered as one Hough vote, the number of false positives can be reduced. However, there are two disadvantages in this approach:
\begin{itemize}
	\item A clustering algorithm is required to approximate the optimal solution to assign voting lines to hypotheses. This clustering algorithm is computationally expensive. To handle this issue, in \cite{VotingLines}, the computational burden of clustering is reduced by reducing the number of extracted image features.
	\item As reported in \cite{VotingLines}, the accuracy of the voting-line based approach is not very high. To obtain a high performance, an extra, pre-trained SVM classifier is used in \cite{VotingLines} to score and verify hypotheses.
\end{itemize}

\subsection{Multi-scale voting scheme} \label{Smsv}

Combining the advantages and disadvantages of both image pyramids and voting lines, we propose a multi-scale voting scheme to handle object scale variations. This scheme inherits the simplicity of image pyramids and avoids detecting objects repeatedly at several scales by integrating the characteristics of voting lines. Furthermore, based on this voting scheme, the correlations between the true and false positives shown in Fig.~\ref{fp:b} can be easily revealed. This leads to a principled and NPMI-based strategy to fuse hypotheses to reduce false positives, which is described in Section~\ref{fusion}.

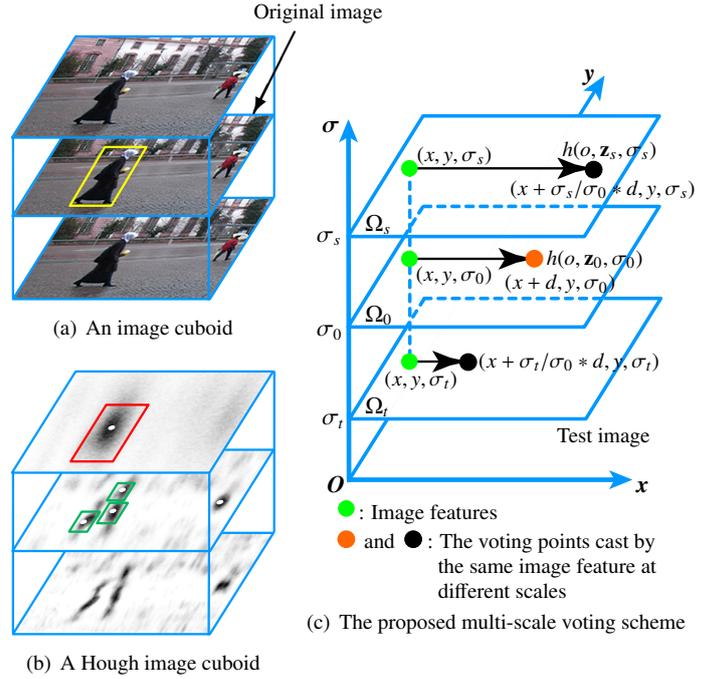
\begin{figure}%
	\centering%
	  \begin{minipage}{0.4\linewidth}
	    \subfigure[An image cuboid]{ \label{msv:a}
	    \begin{overpic}[width=\linewidth]{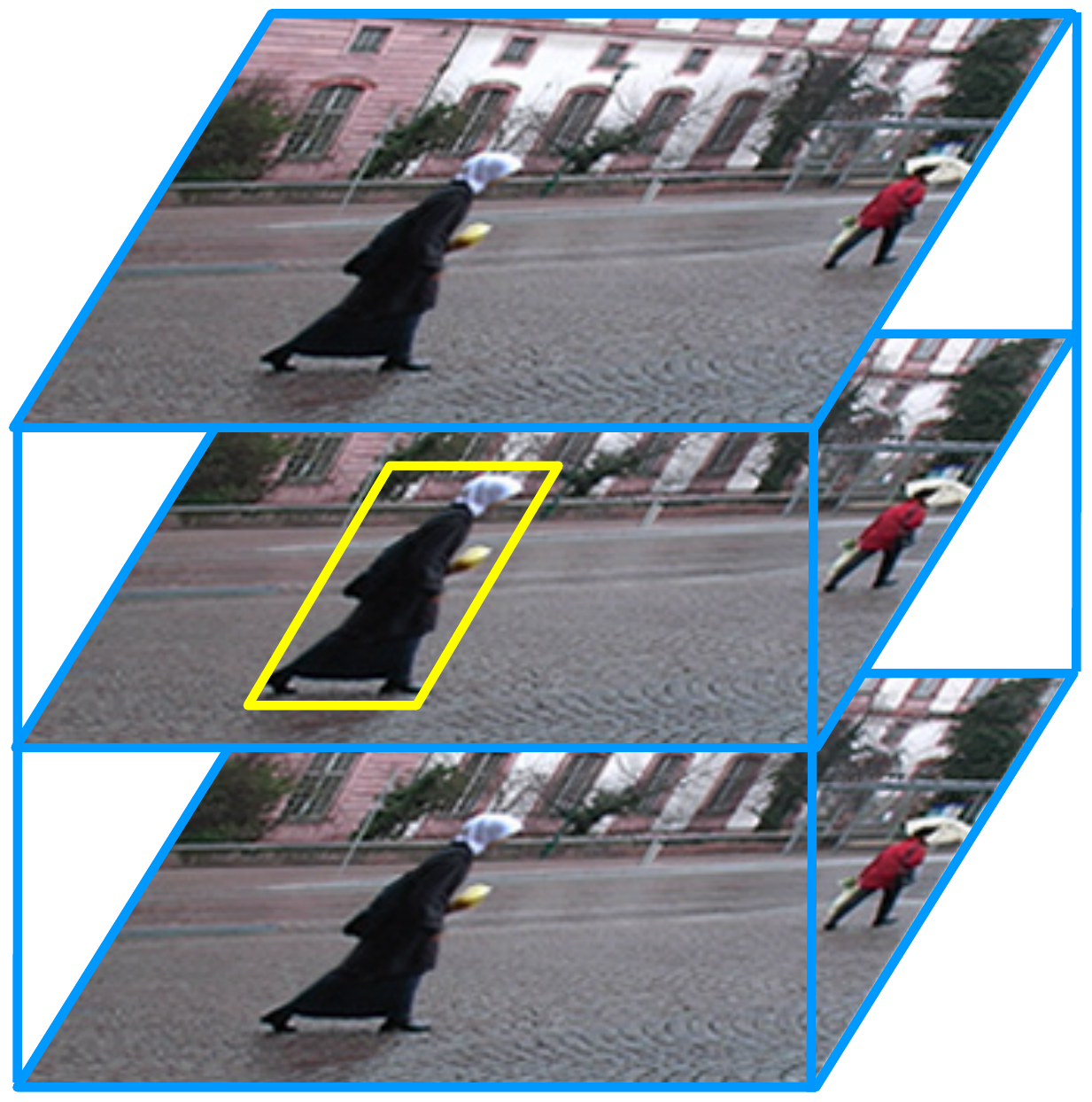}
	    	\put(\LenToUnit{0.91\linewidth},\LenToUnit{1.0617\linewidth}){\footnotesize{Original image}}
	    	\thicklines
	    	\put(\LenToUnit{1.0617\linewidth},\LenToUnit{1.01223\linewidth}){\vector(-1,-2){15}}
	    \end{overpic}
	    }\\
	    \subfigure[A Hough image cuboid]{ \label{msv:b}
	    \includegraphics[width=\linewidth]{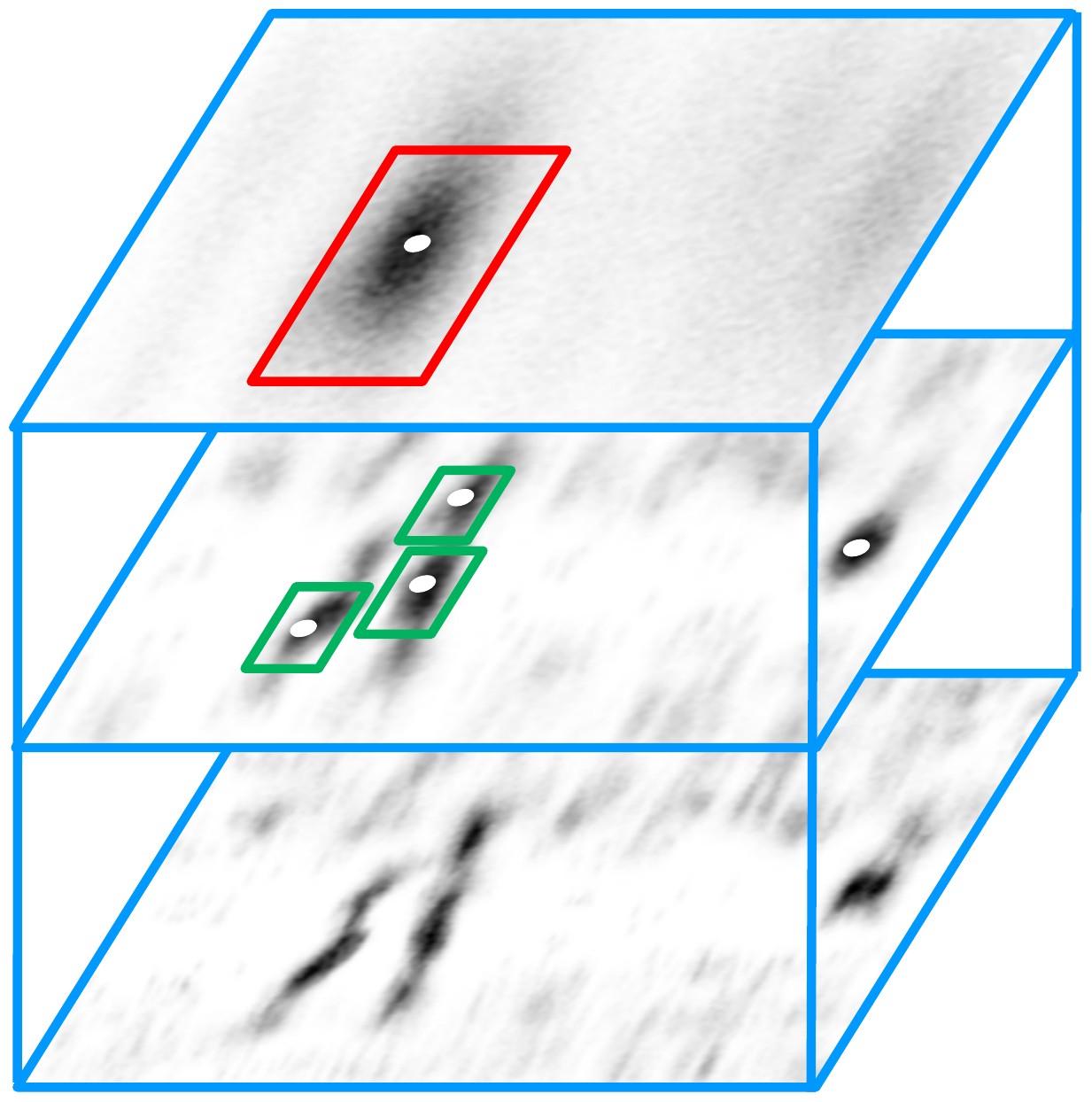}
	    	    }
	  \end{minipage}
	  \hfill
	  \begin{minipage}{0.55\linewidth}
	    \subfigure[The proposed multi-scale voting scheme]{ \label{msv:c}
	    \begin{overpic}[width=\linewidth]{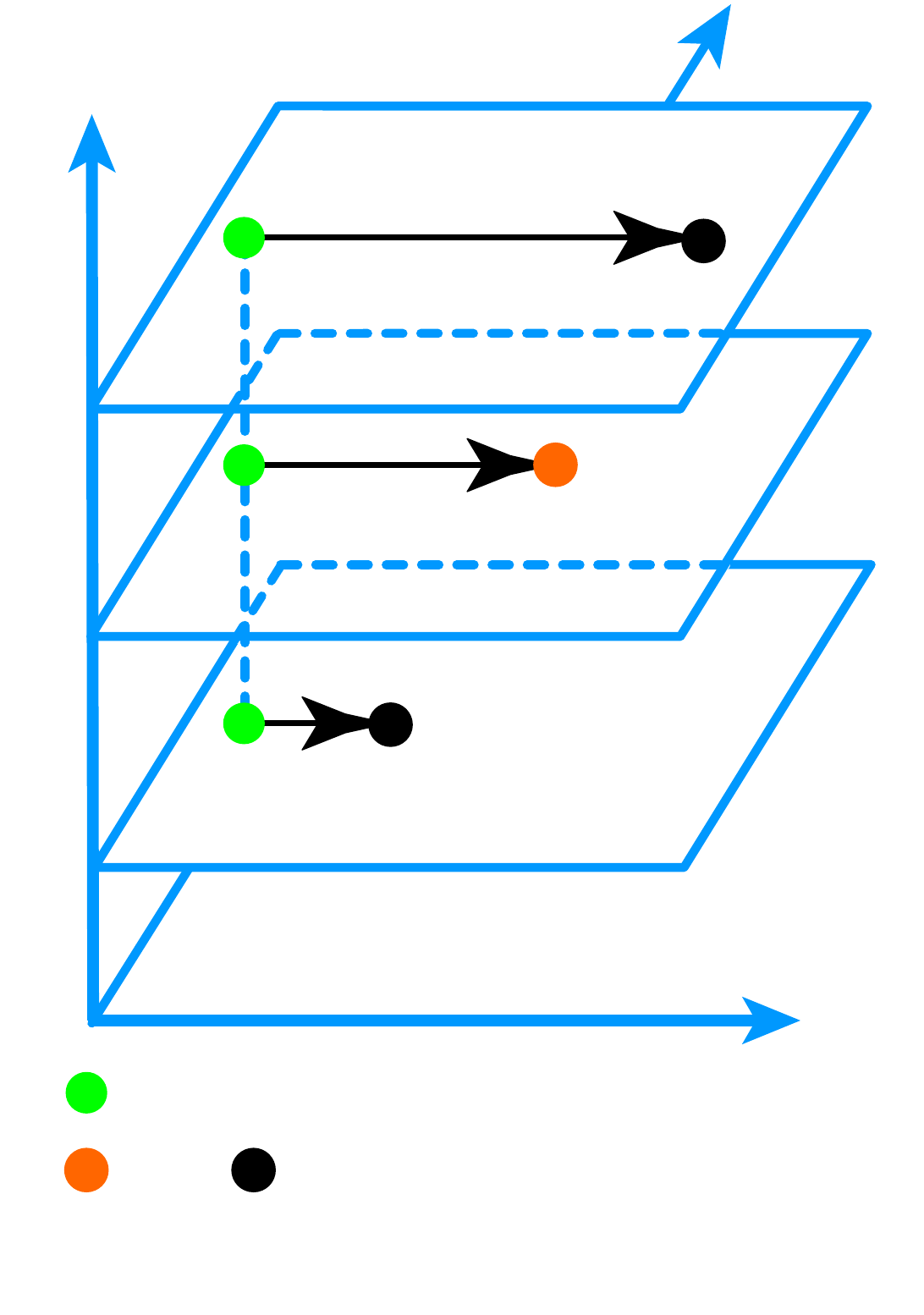}
	    	\put(\LenToUnit{0.028427\linewidth},\LenToUnit{1.265\linewidth}){\small{\bm{$\sigma$}}}
	    	\put(\LenToUnit{0.0142\linewidth},\LenToUnit{0.9665\linewidth}){\small{{$\sigma_s$}}}
	    	\put(\LenToUnit{0.0142\linewidth},\LenToUnit{0.711\linewidth}){\small{{$\sigma_0$}}}
	    	\put(\LenToUnit{0.028427\linewidth},\LenToUnit{0.469\linewidth}){\small{{$\sigma_t$}}}
	    	\put(\LenToUnit{0.042337\linewidth},\LenToUnit{0.2843\linewidth}){\small{\bm{$O$}}}
	    	\put(\LenToUnit{0.145\linewidth},\LenToUnit{0.997\linewidth}){\small{{$\Omega_s$}}}
	    	\put(\LenToUnit{0.145\linewidth},\LenToUnit{0.75\linewidth}){\small{{$\Omega_0$}}}
	    	\put(\LenToUnit{0.145\linewidth},\LenToUnit{0.5\linewidth}){\small{{$\Omega_t$}}}
	    	\put(\LenToUnit{0.197\linewidth},\LenToUnit{0.575\linewidth}){\footnotesize{{$(x,y,\sigma_t)$}}}
	    	\put(\LenToUnit{0.282\linewidth},\LenToUnit{0.860854\linewidth}){\footnotesize{{$(x,y,\sigma_0)$}}}
	    	\put(\LenToUnit{0.2843\linewidth},\LenToUnit{1.18544\linewidth}){\footnotesize{{$(x,y,\sigma_s)$}}}
	    	\put(\LenToUnit{0.4516\linewidth},\LenToUnit{0.625393\linewidth}){\footnotesize{{$(x+\sigma_t/\sigma_0*d,y,\sigma_t)$}}}
	    	\put(\LenToUnit{0.5221573\linewidth},\LenToUnit{0.8375\linewidth}){\footnotesize{{$(x+d,y,\sigma_0)$}}}
	    	\put(\LenToUnit{0.635056\linewidth},\LenToUnit{0.9032\linewidth}){\footnotesize{{$h(o,\mathbf{z}_0,\sigma_0)$}}}
	    	\put(\LenToUnit{0.53627\linewidth},\LenToUnit{1.09444\linewidth}){\footnotesize{{$(x+\sigma_s/\sigma_0*d,y,\sigma_s)$}}}
	    	\put(\LenToUnit{0.6774\linewidth},\LenToUnit{1.19955\linewidth}){\footnotesize{{$h(o,\mathbf{z}_s,\sigma_s)$}}}
	    	\put(\LenToUnit{0.66328\linewidth},\LenToUnit{0.42\linewidth}){\footnotesize{Test image}}
	    	\put(\LenToUnit{0.875\linewidth},\LenToUnit{0.2843\linewidth}){\small{\bm{$x$}}}
	    	\put(\LenToUnit{0.7338427\linewidth},\LenToUnit{1.407\linewidth}){\small{\bm{$y$}}}
	    	\put(\LenToUnit{0.127\linewidth},\LenToUnit{0.2116854\linewidth}){\footnotesize{: Image features}}
	    	\put(\LenToUnit{0.142\linewidth},\LenToUnit{0.132656\linewidth}){\footnotesize{and}}
	    	\put(\LenToUnit{0.31267\linewidth},\LenToUnit{0.06351\linewidth}){\footnotesize{\makecell[l]{: The voting points cast by\\~~the same image feature at\\~~different scales}}}
	    \end{overpic}
	    	}   
	  \end{minipage}
\caption{Illustration of an image cuboid (a), a Hough image cuboid (b) and the proposed multi-scale voting scheme. In (a), the image cuboid is obtained by piling up a few copies of the original image. The Hough image cuboid in (b) is obtained by applying the proposed multi-scale voting scheme on the image cuboid in (a). The red and green boxes in (b) indicate four object hypotheses corresponding to the object in the yellow box in (a). (c) illustrates the voting procedure of the proposed multi-scale voting scheme in a scale space. See text for more details.}
\label{msv}
\end{figure}

We again denote the scales of the objects used in the training stage as $ \sigma_0 $. In the test stage, we first construct an \textit{image cuboid} by piling up a few copies of an original test image as shown in Fig.~\ref{msv:a}. The image cuboid is placed in a scale space as shown in Fig.~\ref{msv:c}. The three levels of the image cuboid are placed on the planes $ \Omega_0$, $ \Omega_s$ and $ \Omega_t$ in the scale space, respectively. The equations of these three planes can be written as $ \sigma=\sigma_0$, $\sigma=\sigma_s$ and $\sigma=\sigma_t$, respectively, where $ \sigma_t < \sigma_0 < \sigma_s $. An image feature is extracted at the coordinates $ (x,y,\sigma_0) $. The same feature can also be found at the coordinates $ (x,y,\sigma_s) $ and $ (x,y,\sigma_t) $. All these three features are represented as three green points in Fig.~\ref{msv:c}. As in Fig.~\ref{vl:a}, by applying a Hough Transform-based detector on the plane $ \Omega_0$, the image feature at the coordinates $ (x,y,\sigma_0) $ may cast a Hough vote to the point at the coordinates $ (x,y+d,\sigma_0) $. Based on the idea of voting lines, without applying a detector on the planes $ \Omega_s$ and $ \Omega_t$, we can directly derive that the image features at the coordinates $ (x,y,\sigma_s) $ and $ (x,y,\sigma_t) $ cast Hough votes to the points at the coordinates $ (x+\sigma_s/\sigma_0*d,y,\sigma_s) $ and $ (x+\sigma_t/\sigma_0*d,y,\sigma_t) $, respectively. In other words, by applying a Hough Transform-based detector only once at any level of an image cuboid, the Hough votes at multiple scales can be obtained simultaneously. We call this voting scheme as a \textit{multi-scale voting} (MSV) scheme.

Now we apply the MSV scheme to the original image in Fig.~\ref{msv:a}. Assume that the original image contains $ r $ image patches $ \{\mathrm{p}_\ell\}_{\ell=1}^r $. The objective is to detect objects at $ S $ different scales denoted as $ \{\sigma_s\}_{s=1}^S $. As explained in Section~\ref{HRM}, each image patch $ \mathrm{p}_\ell $ casts a set of Hough votes: $ \mathcal{E}_\ell = \{\hat{\mathbf{y}}_{\ell_j}\}_{j=0}^m $. Therefore, at each scale $ \sigma_s $, the set of Hough votes cast by all the $ r $ image patches is obtained by:
\begin{equation}
\mathcal{A}^{(\sigma_s)}=\bigcup\nolimits_{n=1}^{r}\mathcal{E}_\ell^{(\sigma_s)} = \bigcup\nolimits_{n=1}^{r} \{\sigma_s/\sigma_0*\hat{\mathbf{y}}_{\ell_j}\}_{j=0}^m,
\end{equation}
where the superscripts indicate the corresponding scale. Thus, we can obtain $ S $ sets of Hough votes $ \{\mathcal{A}^{(\sigma_s)}\}_{s=1}^S $. The corresponding voting points of the $S$ sets of Hough votes form $ S $ Hough images that correspond to the $ S $ scales, respectively. These Hough images form a \textit{Hough image cuboid} (Fig.~\ref{msv:b}).

As with the Hough image pyramid in Fig.~\ref{fp:b}, the Hough image cuboid in Fig.~\ref{msv:b} also contains four hypotheses: one true positive shown in the red box and three false positives shown in the three green boxes. However, since the levels of the image cuboid in Fig.~\ref{msv:a} are identical, the voting points at different levels of the Hough image cuboid are derived from the same original image. In other words, the true positive at the first level and the three false positives at the second level are supported by the same image. Consequently, for any two hypotheses located at different levels of a Hough image cuboid, it is straightforward to reveal and measure the correlation between the two hypotheses by considering the image features which vote for both the hypotheses. Based on this characteristic, we propose an NPMI-based fusion strategy to measure the correlations between hypotheses and reduce false positives in Section~\ref{fusion}. In contrast, for an image pyramid-based approach, object hypotheses obtained at different levels of a Hough image pyramid are supported by different levels of an image pyramid. Thus, it is not straightforward to reveal and measure the correlations between the hypotheses. 

The proposed MSV scheme also possesses some other advantages. First, as mentioned at the beginning of this section, the image cuboid structure used in the MSV scheme inherits the simplicity of image pyramids and is even easier to construct. Second, the more complicated 3D voting lines used in \cite{VotingLines} are not explicitly included. However, the advantages of voting lines are integrated in the proposed MSV scheme. Thus, it is avoided to detect objects repeatedly at several scales, which is required by an image pyramid-based scheme. The feature extraction step and the voting step are performed only once in the proposed MSV scheme. Last but not least, although the MSV scheme is based on the idea of voting lines, the clustering step in \cite{VotingLines} is not required. Rather than using the clustering step, we reduce false positives by using the NPMI-based fusion strategy proposed in Section~\ref{fusion}, which is more simple and more computationally efficient than the clustering step.

\subsection{Probabilistic Framework} \label{pf}
According to the proposed MSV scheme, we propose a probabilistic framework to compute the score of each detection hypothesis. Let $ \mathrm{p}_\ell $ denote an image patch observed at location $ \mathbf{l}_\ell $, and let $h(o,\mathbf{z},\sigma)$ represent a hypothesis of an object category $o$. $h(o,\mathbf{z},\sigma)$ is located at $\mathbf{z}$ at a level of a Hough image cuboid, and the level corresponds to a scale $\sigma$. The probability (i.e., score) of $h(o,\mathbf{z},\sigma)$, $ p(h(o,\mathbf{z},\sigma)) $, means the possibility that $\mathbf{z}$ is the center of an object whose scale is $\sigma$. $ p(h(o,\mathbf{z},\sigma)) $ can be estimated as follows:
\begin{equation}
p(h(o,\mathbf{z},\sigma)) = p(o,\mathbf{z},\sigma) = \sum\limits_\ell  p(o,\mathbf{z},\sigma|\mathrm{p}_\ell,\mathbf{l}_\ell) \  p(\mathrm{p}_\ell,\mathbf{l}_\ell), \label{framework}
\end{equation}
where
\begin{equation}
		p(o,\mathbf{z},\sigma|\mathrm{p}_\ell,\mathbf{l}_\ell)=   p(o|\mathbf{z},\sigma,\mathrm{p}_\ell,\mathbf{l}_\ell) \  p(\mathbf{z},\sigma|\mathrm{p}_\ell,\mathbf{l}_\ell),  \label{equation_2}
\end{equation}%
and $ p(\mathrm{p}_\ell,\mathbf{l}_\ell) $ is assumed to satisfy a uniform distribution since we sample every image patch throughout a test image. The first term on the right side of Eq.~(\ref{equation_2}) indicates the confidence that the image patch $ \mathrm{p}_\ell $ is classified as foreground. This confidence, which can be regarded as the weight assigned to the Hough votes cast by the image patch $ \mathrm{p}_\ell $, is computed as: 
\begin{equation}
  p(o|\mathbf{z},\sigma,\mathrm{p}_\ell,\mathbf{l}_\ell) = p(o|\mathrm{p}_\ell) = \frac{1}{|\mathcal{C}_\ell|}\sum\limits_{\tilde{c} \in \mathcal{C}_\ell} \mathrm{sgn}(\max(\tilde{c},0)), \label{weight}
\end{equation}
where $ \mathrm{sgn}(\cdot) $ is a sign function. The second term on the right side of Eq.~(\ref{equation_2}) specifies the probabilistic Hough vote for the hypothesis $h(o,\mathbf{z},\sigma)$ cast by the image patch $ \mathrm{p}_\ell $, and it is estimated as:
\begin{equation}
	p(\mathbf{z},\sigma|\mathrm{p}_\ell,\mathbf{l}_\ell) 
											= \frac{1}{|\mathcal{E}_\ell|}\sum\limits_{\mathbf{\tilde{e}} \in \mathcal{E}_\ell} \delta_{\mathbf{\tilde{e}}} (\frac{\mathbf{l}_\ell-\mathbf{z}}{\sigma}), \label{vote}
\end{equation}
where $ \delta_{\mathbf{\tilde{e}}} $ is the Dirac function for the voting vector $ \mathbf{\tilde{e}} \in \mathcal{E}_\ell $. 

The probabilities of all possible hypotheses are evaluated. The hypotheses corresponding to the local maxima at each level of a Hough image cuboid are accepted as an initial set of detection results. As shown in Fig.~\ref{msv:b}, an initial set of detection results may contain both true positives and false positives. Therefore, in the next section, we propose a principled and NPMI-based strategy to fuse hypotheses to reduce false positives.

\section{The Fusion of Hypotheses} \label{fusion}

As explained in Section~\ref{multi-voting}, the obtained initial hypotheses may generally contain multiple false positives by using either a Hough image pyramid or a Hough image cuboid. If two hypotheses are obtained at two different levels of a Hough image pyramid, the two hypotheses are derived from \textit{different} levels of an image pyramid (see Subsection~\ref{imagepyra} and Fig.~\ref{fp}). Hence, it is difficult to reveal or measure the correlation between the two hypotheses. In contrast, if two hypotheses locate at two different levels of a Hough image cuboid, the two hypotheses are supported by the \textit{same} test image, because the levels of an image cuboid are identical copies of the original test image. Consequently, the correlation between the two hypotheses can be measured by considering the  common image features shared by the two hypotheses. For two hypotheses at two different scales, a common image feature shared by the two hypotheses means that the image feature votes for both the hypotheses. For example in Fig.~\ref{msv:c}, let us assume that the orange point corresponds to a hypothesis $h(o,\mathbf{z}_0,\sigma_0)$, and the black point on the plane $\Omega_s$ corresponds to another hypotheses $h(o,\mathbf{z}_s,\sigma_s)$. As described in Subsection~\ref{Smsv}, the image cuboid in Fig.~\ref{msv:a} is placed in the scale space in Fig.~\ref{msv:c}, and the three green points in Fig.~\ref{msv:c} represent three identical image features found on the three levels of the image cuboid, respectively. The two identical image features located at the coordinates $(x,y,\sigma_0)$ and $(x,y,\sigma_s)$ vote for the two hypotheses $h(o,\mathbf{z}_0,\sigma_0)$ and $h(o,\mathbf{z}_s,\sigma_s)$, respectively. Since the two image features are identical, we define them as the same common image feature shared by the two hypotheses. For any two hypotheses at two different scales, if more common image features are shared by them, the two hypotheses are considered to be more correlated. 

In this section, we first measure the correlation between any two hypotheses at two different scales by calculating the normalized pointwise mutual information (NPMI) \cite{NPMI} of the two hypotheses. In the calculation of the NPMI between two hypotheses, the common image features shared by the two hypotheses are taken into account by evaluating the contribution from the common image features to the scores of the two hypotheses. The computational burden of evaluating NPMI is less than that of implementing the clustering step in \cite{VotingLines}, and a heuristic threshold is not required when one uses NPMI to judge whether two hypotheses are correlated or not. If two hypotheses at two different scales are considered to be correlated by evaluating NPMI, they are fused to avoid a false positive.

Assume that a Hough image cuboid is derived during the detection of an object category $ o $ on a test image, and the cuboid contains $ S $ levels that correspond to $ S $ scales $ \{\sigma_i\}_{i=1}^S $. All possible hypotheses at scale $ \sigma_i $ form a set of hypotheses denoted as $ \Gamma^{(\sigma_i)}$, and $h(o,\mathbf{z}_i,\sigma_i) \in \Gamma^{(\sigma_i)}$ denotes a hypothesis at scale $\sigma_i$. Similarly, at another scale $ \sigma_j $, a set of hypotheses denoted as $ \Gamma^{(\sigma_j)}$ can be obtained, and $h(o,\mathbf{z}_j,\sigma_j) \in \Gamma^{(\sigma_j)}$ denotes a hypothesis at scale $\sigma_j$. Let $ H^{(\sigma_i)} $ and $ H^{(\sigma_j)} $ be two random variables defined on $ \Gamma^{(\sigma_i)} $ and $ \Gamma^{(\sigma_j)} $, respectively. According to the definition of mutual information proposed in \cite{InfoTheory,InfoTheory2}, the mutual information between $ H^{(\sigma_i)} $ and $ H^{(\sigma_j)} $ is given by:
\begin{equation}
	\begin{split}
	&\mathscr{I}(H^{(\sigma_i)},H^{(\sigma_j)}) \\
	&=\sum\limits_{\scriptsize{\makecell{h(o,\mathbf{z}_i,\sigma_i),\\ h(o,\mathbf{z}_j,\sigma_j)}}} p(h(o,\mathbf{z}_i,\sigma_i),h(o,\mathbf{z}_j,\sigma_j)) \ \ \mathcal{I}(h(o,\mathbf{z}_i,\sigma_i),h(o,\mathbf{z}_j,\sigma_j)),
	\end{split}
\end{equation}
where
\begin{equation}
	\mathcal{I}(h(o,\mathbf{z}_i,\sigma_i),h(o,\mathbf{z}_j,\sigma_j)) = \log \dfrac{p(h(o,\mathbf{z}_i,\sigma_i),h(o,\mathbf{z}_j,\sigma_j))}{p(h(o,\mathbf{z}_i,\sigma_i))\ p(h(o,\mathbf{z}_j,\sigma_j))}.
\end{equation}
In \cite{NPMI}, $\mathcal{I}(h(o,\mathbf{z}_i,\sigma_i),h(o,\mathbf{z}_j,\sigma_j))$ is defined as the pointwise mutual information (PMI) between a pair of hypotheses $ (h(o,\mathbf{z}_i,\sigma_i),h(o,\mathbf{z}_j,\sigma_j)) $. The values of PMI in the following three typical situations are determined as follows:
\begin{itemize}
	\item When hypotheses $ h(o,\mathbf{z}_i,\sigma_i) $ and $ h(o,\mathbf{z}_j,\sigma_j) $ are supported by the same image features, the two hypotheses are completely correlated and definitely correspond to the same object. In this situation, we have 
	\begin{equation}
		\begin{split}
		p(h(o,\mathbf{z}_i,\sigma_i),h(o,\mathbf{z}_j,\sigma_j)) &= p(h(o,\mathbf{z}_i,\sigma_i))\\
		                                       &= p(h(o,\mathbf{z}_j,\sigma_j)),
		\end{split}
	\end{equation}
	\begin{equation}
		\mathcal{I}(h(o,\mathbf{z}_i,\sigma_i),h(o,\mathbf{z}_j,\sigma_j)) = -\log p(h(o,\mathbf{z}_i,\sigma_i),h(o,\mathbf{z}_j,\sigma_j)).
	\end{equation}
	\item When hypotheses $ h(o,\mathbf{z}_i,\sigma_i) $ and $ h(o,\mathbf{z}_j,\sigma_j) $ are statistically independent, we have
	\begin{equation}
			p(h(o,\mathbf{z}_i,\sigma_i),h(o,\mathbf{z}_j,\sigma_j)) = p(h(o,\mathbf{z}_i,\sigma_i))\ p(h(o,\mathbf{z}_j,\sigma_j)),
	\end{equation}
	\begin{equation}
			\mathcal{I}(h(o,\mathbf{z}_i,\sigma_i),h(o,\mathbf{z}_j,\sigma_j)) = 0.
	\end{equation}
	In this situation, the two hypotheses are completely uncorrelated. They are not supported by the same image features, but they still share some common image features. 
	\item When hypotheses $ h(o,\mathbf{z}_i,\sigma_i) $ and $ h(o,\mathbf{z}_j,\sigma_j) $ are supported by completely different image features, the two hypotheses definitely correspond to different objects. In this situation, we have
	\begin{equation}
		p(h(o,\mathbf{z}_i,\sigma_i),h(o,\mathbf{z}_j,\sigma_j)) = 0,
	\end{equation}
	\begin{equation}
		\mathcal{I}(h(o,\mathbf{z}_i,\sigma_i),h(o,\mathbf{z}_j,\sigma_j)) = -\infty.
	\end{equation}
\end{itemize}

Therefore, the value of PMI is bounded in the interval $ [-\infty,-\log p(h(o,\mathbf{z}_i,\sigma_i),h(o,\mathbf{z}_j,\sigma_j))] $. However, the interval is not symmetric about zero, the upper bound is not fixed, and the lower bound approaches infinity. PMI is normalized in \cite{NPMI} to obtain fixed bounds that are symmetric about zero as follows:

	\begin{align}
	&\mathcal{I}_n(h(o,\mathbf{z}_i,\sigma_i),h(o,\mathbf{z}_j,\sigma_j))  \notag \\ 
	&= \dfrac{\quad \log \dfrac{p(h(o,\mathbf{z}_i,\sigma_i),h(o,\mathbf{z}_j,\sigma_j))}{p(h(o,\mathbf{z}_i,\sigma_i))\ p(h(o,\mathbf{z}_j,\sigma_j))}\quad} {-\log p(h(o,\mathbf{z}_i,\sigma_i),h(o,\mathbf{z}_j,\sigma_j))} \notag \\ 
	 &= \dfrac {\log\dfrac{p(h(o,\mathbf{z}_j,\sigma_j)|h(o,\mathbf{z}_i,\sigma_i))}{p(h(o,\mathbf{z}_j,\sigma_j))}} {\quad -\log [p(h(o,\mathbf{z}_i,\sigma_i)) \  p(h(o,\mathbf{z}_j,\sigma_j)|h(o,\mathbf{z}_i,\sigma_i))] \quad}. \label{NPMI}
	 \end{align}

$ \mathcal{I}_n(h(o,\mathbf{z}_i,\sigma_i),h(o,\mathbf{z}_j,\sigma_j)) $ is called the normalized pointwise mutual information between a pair of hypotheses $ (h(o,\mathbf{z}_i,\sigma_i),h(o,\mathbf{z}_j,\sigma_j)) $. The values of NPMI in the above-mentioned three typical situations are $ 1 $, $ 0 $ and $ -1 $, respectively. Hence, the value of NPMI is bounded in the interval $ [-1,1] $. With the fixed and symmetric bounds, in practice, calculating NPMI is more flexible than calculating PMI. Moreover, the correlations between different pairs of hypotheses are comparable.

In Eq.~(\ref{NPMI}), given that $ h(o,\mathbf{z}_i,\sigma_i) $ corresponds to an object, $ p(h(o,\mathbf{z}_j,\sigma_j)|h(o,\mathbf{z}_i,\sigma_i)) $ means the probability that $ h(o,\mathbf{z}_j,\sigma_j) $ corresponds to the same object. $ p(h(o,\mathbf{z}_i,\sigma_i)) $ and $ p(h(o,\mathbf{z}_j,\sigma_j)) $ in Eq.~(\ref{NPMI}) can be calculated by using Eq.~(\ref{framework}). We estimate $ p(h(o,\mathbf{z}_j,\sigma_j)|h(o,\mathbf{z}_i,\sigma_i)) $ by using a kernel density estimation technique \cite{KDE}:
\begin{equation}
	\begin{split}
	&p(h(o,\mathbf{z}_j,\sigma_j)|h(o,\mathbf{z}_i,\sigma_i))\\ &\propto\frac{1}{b^2 \sum_\ell w(\mathrm{p}_\ell)}\sum_\ell \mathcal{K}\Big[\dfrac{\frac{\sigma_j}{\sigma_i} \cdot (\mathbf{z}_i-\mathbf{l}_\ell)+\mathbf{l}_\ell-\mathbf{z}_j}{b}\Big] w(\mathrm{p}_\ell), \label{hjhi}
	\end{split}
\end{equation}
where $ w(\mathrm{p}_\ell) = p(o,\mathbf{z}_i,\sigma_i|\mathrm{p}_\ell,\mathbf{l}_\ell) $ is defined in Eq.~(\ref{equation_2}); $ \mathcal{K} $ is a nonnegative, radially symmetric kernel function; $ b $ is the kernel bandwidth; $ \mathbf{l}_\ell $ is the location of image patch $ \mathrm{p}_\ell $.

When an initial set of detection results is obtained as explained in Subsection~\ref{pf}, the NPMI for each pair of hypotheses $ (h(o,\mathbf{z}_i,\sigma_i),h(o,\mathbf{z}_j,\sigma_j))(i \neq j) $ is evaluated. If $ \mathcal{I}_n(h(o,\mathbf{z}_i,\sigma_i),h(o,\mathbf{z}_j,\sigma_j))>0 $, $ h(o,\mathbf{z}_i,\sigma_i) $ and $ h(o,\mathbf{z}_j,\sigma_j) $ are statistically dependent. In this case, $ h(o,\mathbf{z}_i,\sigma_i) $ and $ h(o,\mathbf{z}_j,\sigma_j) $ are fused together. If $ p(h(o,\mathbf{z}_i,\sigma_i)) > p(h(o,\mathbf{z}_j,\sigma_j)) $, $ h(o,\mathbf{z}_i,\sigma_i) $ is reserved as the fusion result, while $ h(o,\mathbf{z}_j,\sigma_j) $ is removed.

\section{Experiments} \label{experiments}

The proposed HRM approach is evaluated on challenging datasets and compared with several state-of-the-art approaches. {Furthermore, four variants of the HRM approach are also implemented and evaluated on the challenging datasets. These variants are compared with the HRM approach to evaluate the influences of its components on its performance.}

\subsection{Experimental settings}
The experimental settings for the HRM approach and all of its variants are as follows. For training, we randomly sample 12,000 positive and 12,000 negative training image patches in each experiment. Context information is extracted for each image patch by using $ 16 $ neighboring patches (as shown in Fig.~\ref{Context}). Thus, the numbers of HRMs and LRMs used in the experiments are both 17. The size of each image patch is $ 16 \times 16 $ pixels. As in \cite{HF09}, we use 13 feature channels to obtain feature vectors, which include the absolute values of the two components of the gradient operator, the absolute values of the two components of the Laplacian operator and the nine channels of the HOG feature \cite{HOG}. A $ 5 \times 5 $ region centered at each pixel is used to extract the HOG feature. The min and max filters are applied on the 13 channels to yield 26 feature channels. The feature vector of each image patch is obtained by concatenating the 26 feature channels of each pixel in the image patch. In BPLS and PLS, the number of latent components is determined to be $ 100 $ by using a cross-validation procedure. As for the ridge-parameter $ \alpha $ used in BPLS, it is chosen to be $ 10^{-10} $ as suggested in \cite{BridgePLS}. In the experiments, we also tried other values of $ \alpha $. As stated in \cite{BridgePLS}, the performance of BPLS greatly approximates to that of PLS when $ \alpha $ is very small (e.g., $ 10^{-7} $, $ 10^{-10} $, etc.), and degrades when $ \alpha $ is close to $ 1 $. For each test image, we densely sample image patches from it and generate a 4-level Hough image cuboid (or a 4-level Hough image pyramid) corresponding to 4 scales (i.e., 0.75, 1, 1.25 and 1.5).

{The proposed HRM approach mainly consists of three components, i.e., a BPLS-based module for computing HRMs, an MSV scheme for handling object scale variations and an NPMI-based strategy for fusing hypotheses. The influences of the three components on the performance of the HRM approach are evaluated by comparing the HRM approach with its four variants. The four variants are the PSCG \cite{PSCG}, HRM1, HRM2 and HRM3 approaches, which are listed in Table~\ref{variants}. The PSCG approach is the early version of the HRM approach. The second column in Table~\ref{variants} indicates the techniques used to compute HRMs in the variants. The third column in Table~\ref{variants} shows the schemes used to handle object scale variations in the variants, in which \textquotesingle image pyramid\textquotesingle~means that object scale variations are handled by performing object detection at each level of an image pyramid separately. The fourth column in Table~\ref{variants} gives the strategies used to fuse hypotheses in the variants. In the fourth column, \textquotesingle Non-maxima suppression\textquotesingle~(NMS) means the hypotheses corresponding to the 3D local maxima in a Hough image pyramid or a Hough image cuboid are accepted as the final detection result. Essentially, the commonly used NMS strategy is also a kind of fusion strategy.

\begin{table}
\caption{{The components of the HRM approach and its four variants. See text for more details.}}\label{variants}
\centering
\begin{tabular}{|c|c|c|c|} 
    \hline 
      \textbf{Name} & \textbf{\makecell{Computing\\ HRMs}} & \textbf{\makecell{Handling object\\ scale variations}} & \textbf{\makecell{Fusion of\\ hypotheses}}\\
    \Xhline{1.2pt}
      \makecell{PSCG\\ \cite{PSCG}} & PLS & Image pyramid & \makecell{Non-maxima\\ suppression}\\
    \hline
    	HRM-1 & BPLS & Image pyramid & \makecell{Non-maxima\\ suppression}\\
    \hline
    	HRM-2 & BPLS & MSV & \makecell{Non-maxima\\ suppression}\\
    \hline
    	\multirow{2}{*}{HRM-3} & \multirow{2}{*}{PLS} & \multirow{2}{*}{MSV} & \multirow{2}{*}{NPMI} \\
    	&&&\\
    \hline
    	\multirow{2}{*}{HRM} & \multirow{2}{*}{BPLS} & \multirow{2}{*}{MSV} & \multirow{2}{*}{NPMI} \\
    	&&&\\
    \hline
\end{tabular}
\end{table}%

The PSCG approach uses traditional PLS technique, and the image pyramid-based scheme and the NMS strategy employed in the PSCG approach are both commonly used. Therefore, we use the PSCG approach as a baseline in the comparison among the HRM approach and its variants. The difference between the performances of PLS and BPLS can be evaluated by comparing the PSCG and HRM-1 approaches. The difference can also be indicated by comparing the HRM-3 and HRM approaches. The comparison between the HRM-1 and HRM-2 approaches can evaluate the improvement obtained by using the proposed MSV scheme instead of the image pyramid-based scheme. By comparing the HRM-2 and HRM approaches, we can evaluate the improved performance of the proposed NPMI-based strategy compared with that of the NMS strategy.}

\subsection{Experimental Results}

\begin{figure}
    \centering 
    \includegraphics[scale=0.171]{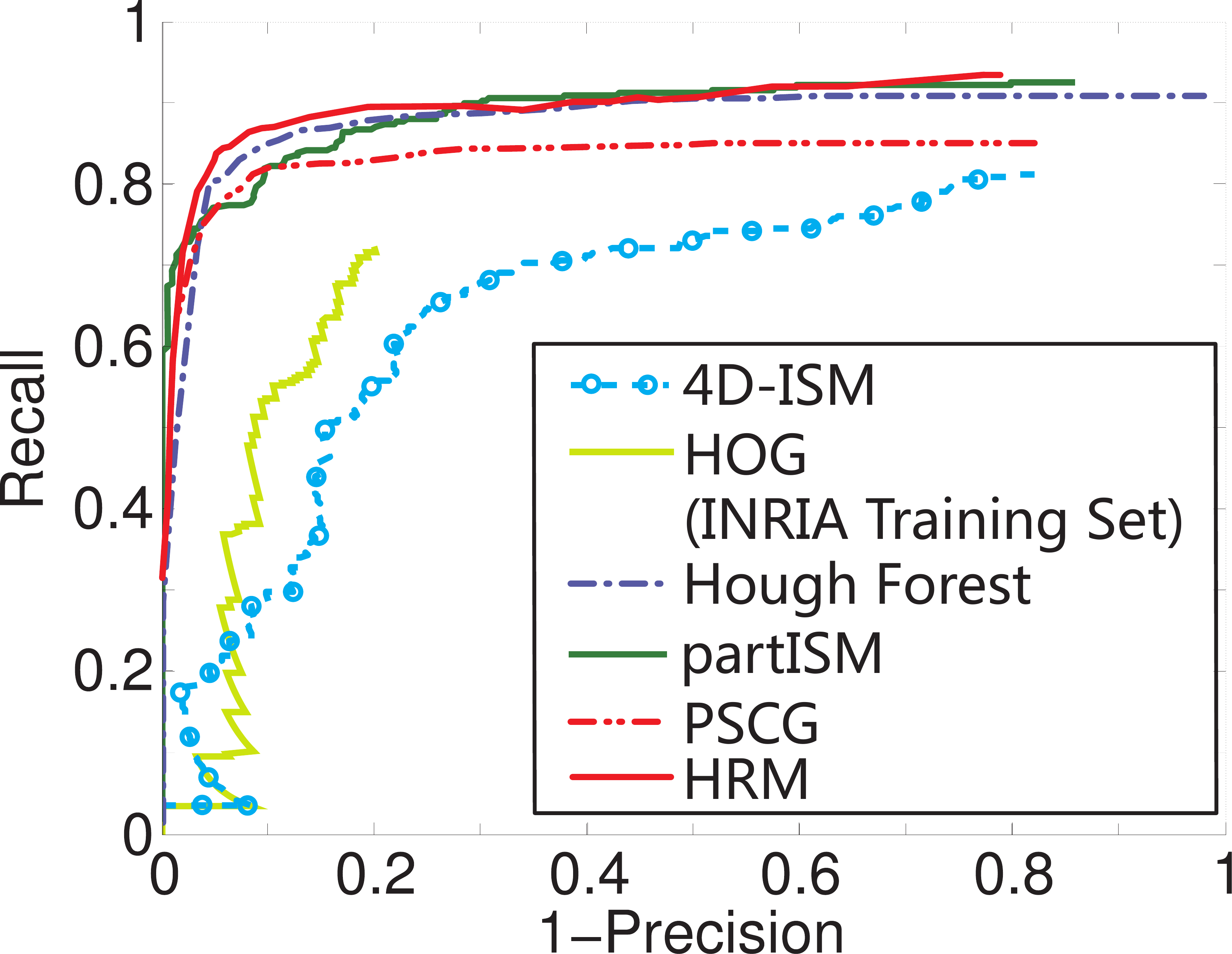}
  \caption{Comparison of the proposed HRM approach and other competing approaches on the TUD Pedestrians dataset. We use some results of the competing approaches given in \cite{partISM, HF09}. Note that 4D-ISM \cite{4dism} uses only 210 training images provided in the dataset.}  \label{Curve}
\end{figure}

{\bfseries The TUD Pedestrians} \cite{partISM} is a challenging pedestrian dataset containing partial occlusions, cluttered background, and dramatic scale changes. We use all the provided 400 training images and 250 test images that contain 311 pedestrians in this dataset. The obtained precision-recall curves are given in Fig.~\ref{Curve}, which shows that the HRM approach significantly outperforms the 4D-ISM \cite{4dism} and HOG \cite{HOG} approaches. Moreover, in Fig.~\ref{Curve}, although the recall rate of the HRM approach is slightly lower than that of the partISM \cite{partISM} approach when the precision rate is between $47\%$ and $72\%$, the overall performance of the HRM approach is better than those of the partISM and Hough forest \cite{HF09} approaches.

\begin{table}
\caption{The EERs obtained by the eight competing approaches on the TUD Pedestrians dataset. The HOG approach \cite{HOG} is not listed because it stops at a recall of 60\% and a precision of 75\% \cite{4dism}, as can be seen in Fig.~\ref{Curve}.}\label{pede}
\centering
\begin{tabular}{| p{0.6\linewidth}<{\centering} | p{0.3\linewidth}<{\centering} |} 
    \hline 
      \textbf{Approach} \hfill & \textbf{EER} \\
    \Xhline{1.2pt}
      4D-ISM \cite{4dism} & 68\%\\
    \hline
      partISM \cite{partISM} & 84\%\\
    \hline
      Hough forest \cite{HF09} & 86.5\%\\
    \hline
      PSCG \cite{PSCG}& 83.0\%\\
    \hline
      \textbf{HRM-1}& \textbf{83.4\%}\\
	\hline
      \textbf{HRM-2}& \textbf{84.5\%}\\  
    \hline
      \textbf{HRM-3}& \textbf{88.4\%}\\    
    \hline
      \textbf{HRM}& \textbf{87.7\%}\\
    \hline
\end{tabular}
\end{table}%

\begin{figure}
    \centering 
    \includegraphics[scale=0.171]{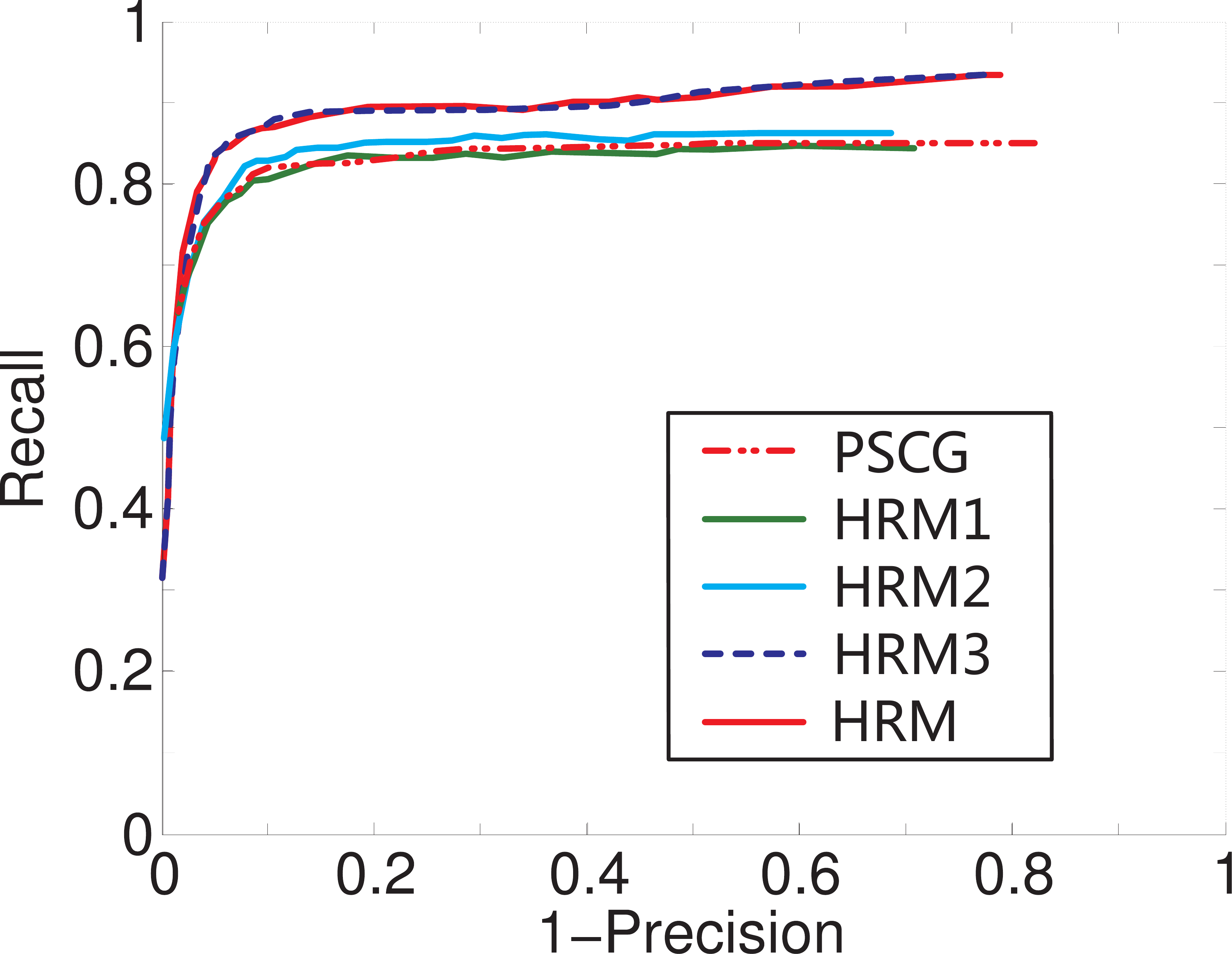}
  \caption{{Comparison among the HRM approach and its four variants on the TUD Pedestrians dataset. All the five approaches use the 400 training images provided in the dataset}.}  \label{Curve-mine}
\end{figure}

{The precision-recall curves of the HRM approach and its variants are given in Fig.~{\ref{Curve-mine}}. It can be seen that the HRM and HRM-3 approaches achieve very similar performances and outperform other variants. Moreover, the PSCG and HRM-1 approaches also obtain very similar performances, which are slightly inferior to the HRM-2 approach.}

The Equal Error Rates (EERs) obtained by all the above-mentioned competing approaches on the TUD Pedestrians dataset are listed in Table~{\ref{pede}}. The HRM-3 approach obtains the highest EER of $ 88.4\% $, while the second highest EER of $ 87.7\% $ is obtained by the HRM approach. {It can be seen from Fig.~{\ref{Curve-mine}} and Table~{\ref{pede}} that the performance of the HRM approach is very close to that of the HRM-3 approach, and the PSCG and HRM-1 approaches also obtain similar performances. Therefore, the effectiveness of BPLS and PLS is very similar. However, in the training stage of the experiments, BPLS is about 10 times faster than PLS in average. This shows that BPLS can significantly improve the efficiency of the HRM approach without reducing its effectiveness.}

{\bfseries The TUD Motorbikes} dataset \cite{TUDMotor} contains 125 side-views of motorbikes in 115 images. This dataset includes challenging scale and illumination changes, partial occlusions, and cluttered background. We choose 284 images captured in real scenes from the Caltech motorbikes dataset \cite{Motor_train} for training. Table \ref{motor} shows the EERs obtained by nine approaches. {Both the HRM approach and the HRM-3 approach achieve the EER of $90.0 \%$ which shows that they outperform all the competing approaches. The HRM-2 approach obtains an inferior performance when it is compared with the Boosted Random Ferns {\cite{Boosted_Random_Ferns}} and  ISM+MDL {\cite{ISM08}} approaches, but it outperforms the HRM-1 and PSCG approaches. Again, the HRM-1 and PSCG approaches obtain similar performances, which are close to that of the Fast PRISM approach {\cite{FastPRISM}} and are superior to that of the IRD approach {\cite{TUDMotor}}.}

\begin{table} 
\caption{The EERs obtained by the nine competing approaches on the TUD Motorbikes dataset.}\label{motor}
\centering
\begin{tabular}{| p{0.6\linewidth}<{\centering} | p{0.3\linewidth}<{\centering} |} 
    \hline 
      \textbf{Approach} \hfill & \textbf{EER} \\
    \Xhline{1.2pt}
      IRD \cite{TUDMotor} & 81.0\%\\
    \hline
      ISM+MDL \cite{ISM08} & 87.0\%\\
    \hline
      Fast PRISM \cite{FastPRISM} & 83.0\%\\
    \hline
      Boosted Random Ferns \cite{Boosted_Random_Ferns} & 89.3\%\\
    \hline
      PSCG \cite{PSCG}& 82.7\%\\
    \hline
      \textbf{HRM-1}& \textbf{82.9\%}\\
    \hline
          \textbf{HRM-2}& \textbf{84.6\%}\\
    \hline
          \textbf{HRM-3}& \textbf{90.0\%}\\
    \hline
          \textbf{HRM}& \textbf{90.0\%}\\
    \hline
\end{tabular}
\end{table}%

\subsection{Experimental Analysis}

\begin{figure*}[t]
  \centering 
  \includegraphics[width=0.9\linewidth]{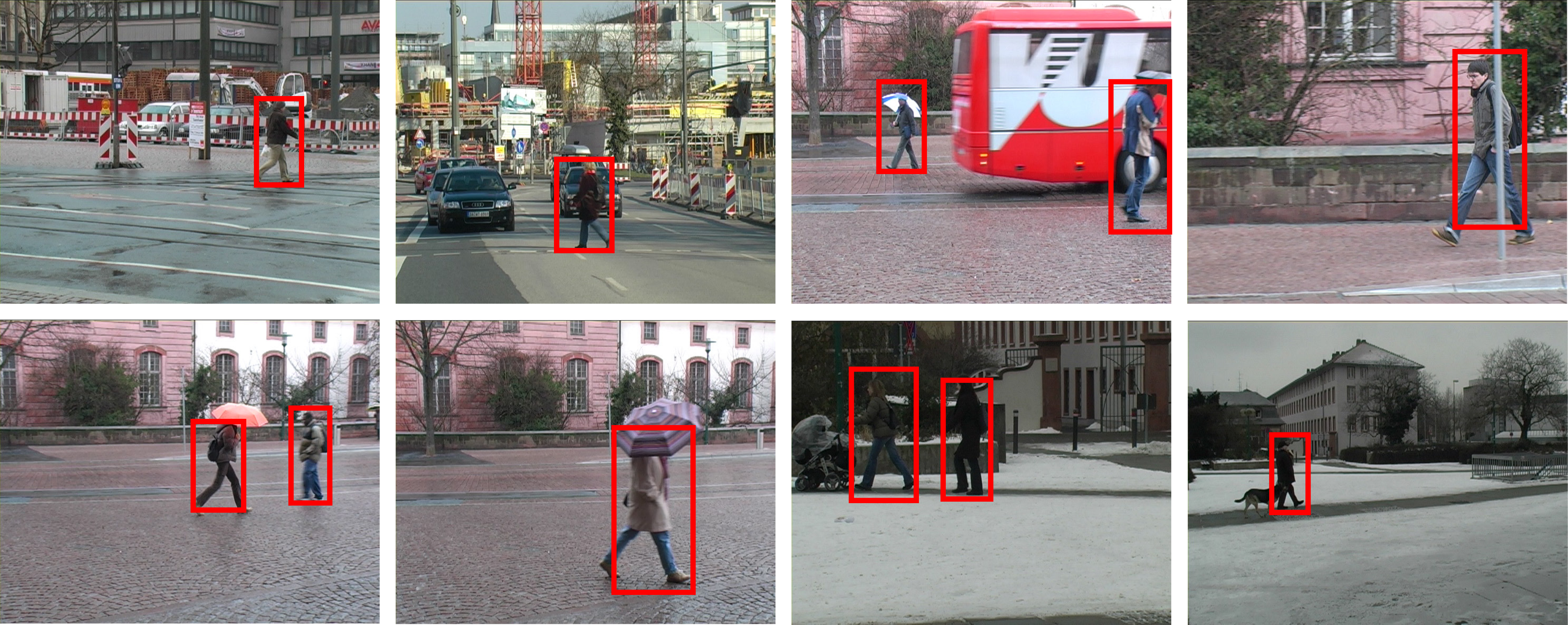}
  \caption{Examples of some results obtained by the proposed HRM approach on the TUD Pedestrians dataset.}  \label{result_pede}
\end{figure*}

\begin{figure*}[t]
  \centering 
  \includegraphics[width=0.885\linewidth]{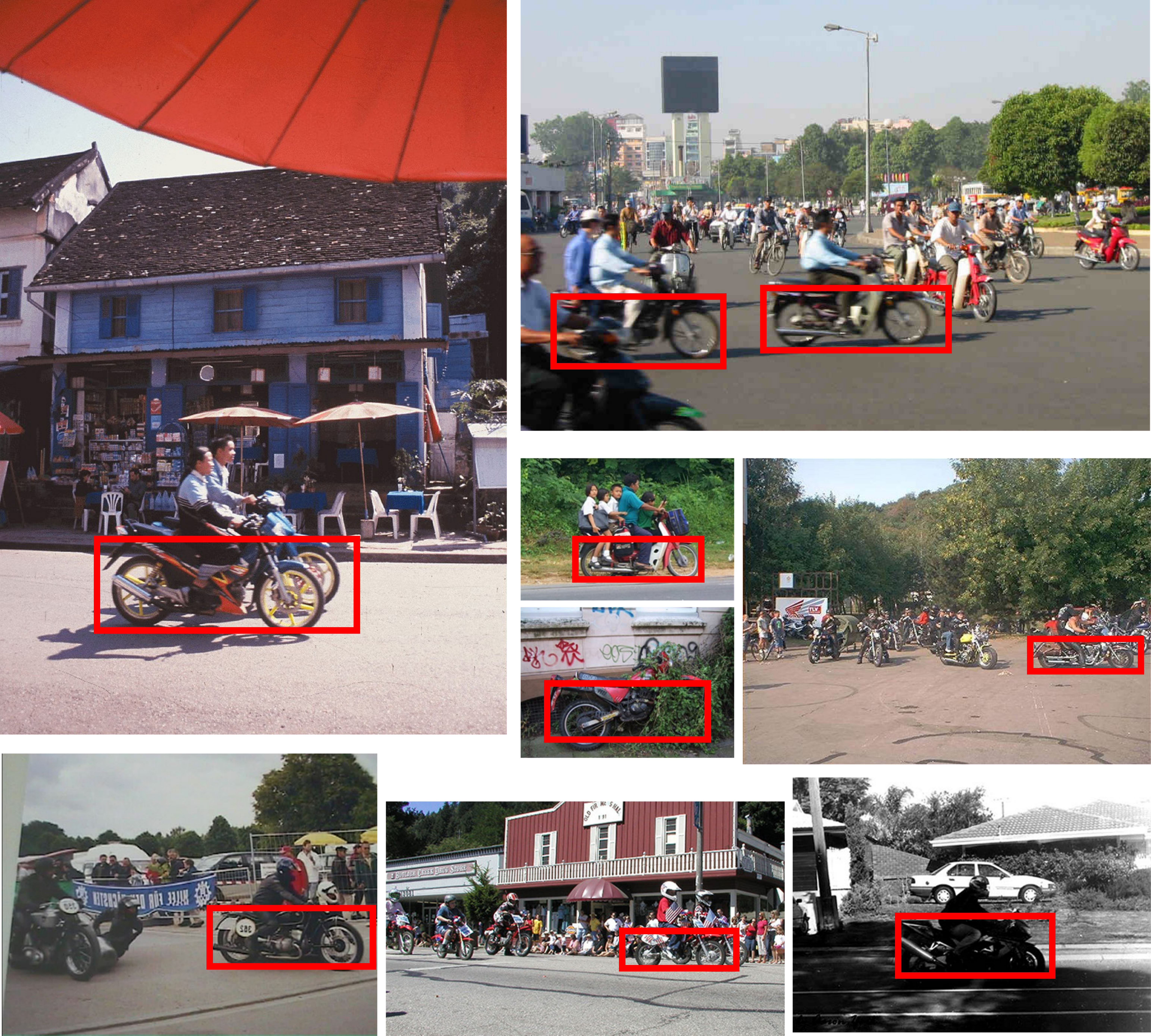}
  \caption{Examples of some results obtained by the proposed HRM approach on the TUD Motorbikes dataset.}  \label{result_motor}
\end{figure*}

In the above experiments, except for the Boosted Random Ferns approach, all other competing approaches are classical Hough Transform-based approaches. From the experimental results we can see that the derived HRMs can effectively model the relationship between context-encoded feature vectors and Hough votes, and the proposed MSV scheme and NPMI-based fusion strategy can effectively deal with object scale changes. Object locations can be accurately predicted by the HRM approach against partial occlusions, cluttered background, dramatic scale variations and drastic illumination changes (see Figs.~\ref{result_pede} and \ref{result_motor} for some examples).

{With regard to the components of the HRM approach, we analyze their influences on the performance of the HRM approach based on the above-mentioned experimental results (as shown in Fig.~\ref{Curve-mine}, Table~\ref{pede} and Table~\ref{motor}).
\begin{itemize}
	\item On both the TUD Pedestrians and TUD Motorbikes datasets, the HRM and HRM-3 approaches obtain very similar performances. Moreover, the HRM-1 and PSCG approaches also achieve similar performances on both the datasets. These results indicate that the accuracy of BPLS is quite close to that of PLS. Hence, introducing the ridge-parameter $ \alpha $ and extracting all latent components simultaneously in one eigenvalue decomposition step in BPLS do not degrade the performance as compared to PLS. Furthermore, in the experiments, BPLS is about 10 times faster than PLS in computing HRMs. Therefore, by using BPLS, we can significantly improve the efficiency of the training stage of the HRM approach while keeping the accuracy in the detection results. 
	
	\item The HRM-2 approach achieves slightly better performance as compared to the PSCG and HRM-1 approaches on both the datasets. This suggests that, even without using the proposed NPMI-based fusion strategy, the proposed MSV scheme can still perform better than an image pyramid-based scheme and slightly improve the accuracy of the detection results. On the  TUD Pedestrians and TUD Motorbikes datasets, the MSV scheme helps the HRM-2 approach to obtain the improvements of $1.1\%$ and $1.7\%$ in EER over the HRM-1 approach, respectively. Consequently, the voting vectors produced by using the MSV scheme are more accurate than those produced by using an image pyramid-based scheme. Note that the main purpose of the MSV scheme is not to improve the performance but to serve as a foundation for revealing the correlations between hypotheses at different scales (see Section~\ref{multi-voting}).
	
	\item The performances of the HRM and HRM-3 approach are obviously superior to that of the HRM-2 approach on both the datasets. This shows that the proposed NPMI-based fusion strategy used in the HRM approach is more effective than the commonly used NMS strategy employed in the HRM-2 approach. The NPMI-based fusion strategy can effectively reduce false positives and improve the performance of the HRM approach. On the  TUD Pedestrians and TUD Motorbikes datasets, the NPMI-based fusion strategy helps the HRM approach to obtain the improvements of $3.2\%$ and $5.4\%$ in EER over the HRM-2 approach, respectively.
	
	\item {The NPMI-based fusion strategy is not only effective but also relatively simple. It requires only simple calculations as shown in Eqs.~(\ref{NPMI}) and (\ref{hjhi}). In contrast, the ISM+MDL \cite{ISM08} and IRD \cite{TUDMotor} approaches require more complicated verification steps, i.e., segmentation and SVM-based classification, respectively. Moreover, the performances of these two approaches are inferior to that of the HRM approach on the TUD Motorbikes dataset}.
\end{itemize}
}
\section{Conclusion} \label{conclusion}

In this paper, we have proposed a novel Hough Transform-based object detection approach named the HRM approach. We have employed Bridge Partial Least Squares to efficiently establish context-encoded Hough Regression Models. PLS can reduce the redundancy and eliminate the multicollinearity of a feature set. The only parameter used in PLS is the number of latent components, which can be determined by using a cross-validation procedure. BPLS is an efficient variant of PLS, which can simultaneously extract all latent components for feature vectors by using eigenvalue decomposition only once. Due to these advantages of BPLS, the obtained HRMs can accurately generate Hough votes for possible object locations. Moreover, we have proposed a novel multi-scale voting scheme to efficiently handle object scale changes. This scheme casts Hough votes at multiple scales simultaneously by using only an original image. Therefore, constructing an image pyramid and detecting objects repeatedly on all levels of the image pyramid as in many other approaches are not required. Based on this scheme, normalized pointwise mutual information between estimated hypotheses can be evaluated, and it is used to fuse multiple hypotheses corresponding to the same object to reduce false positives. In the experiments, we have also compared the HRM approach with its four variants to evaluate the influences of its components on its performance. Experimental results have shown that the proposed approaches have achieved better performances than several state-of-the-art competing approaches on challenging datasets.%

\section*{Acknowledgement}

This work was supported by the National Natural Science Foundation of China under Grants 61170179, 61472334 and 61201359, by the Natural Science Foundation of Fujian Province of China under Grant 2012J05126, by the Specialized Research Fund for the Doctoral Program of Higher Education of China under Grant 20110121110033.

\bibliographystyle{bibliographystyle(model1-num-names)}
\bibliography{references}

\parpic{\includegraphics[width=1in,height=1.25in,clip]{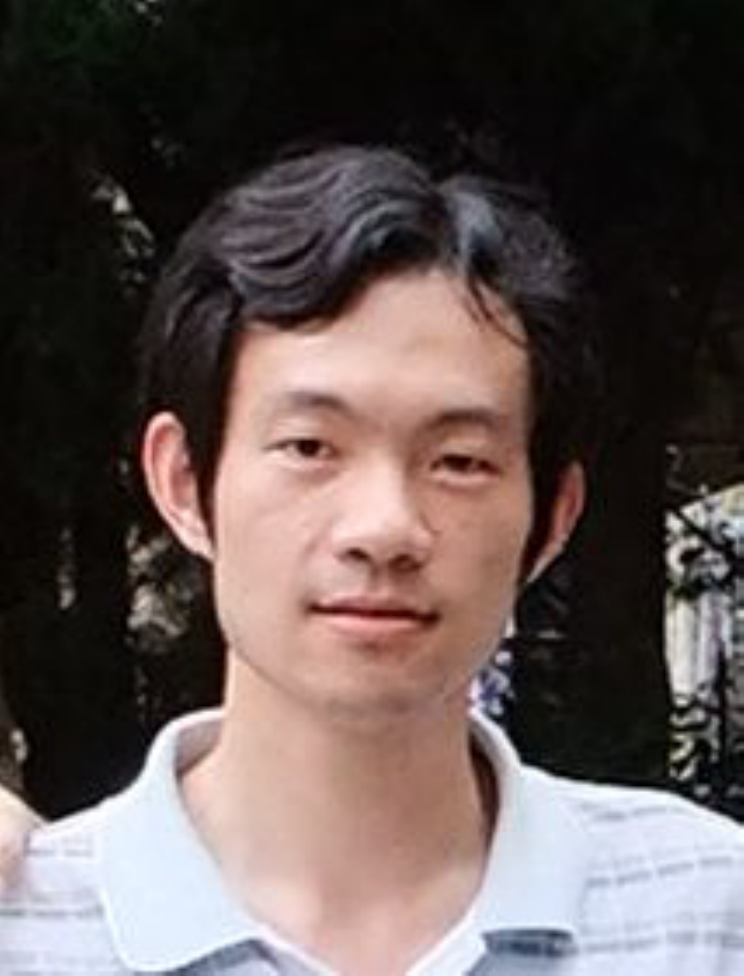}}
\noindent {\bf Jianyu Tang} received his B.S. degree in Computer Science from Wuhan University of Science and Technology, Wuhan, China, in 2002 and the M.E. degree in Software Engineering from Wuhan University, Wuhan, China, in 2008. He is currently a PhD student in the School of Information Science and Technology at Xiamen University, Xiamen, China. His research interests include visual object detection, visual tracking and machine learning.

\parpic{\includegraphics[width=1in,height=1.25in,clip]{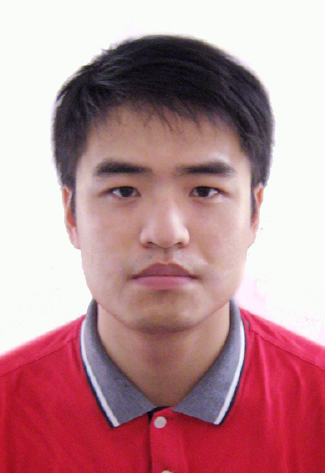}}
\noindent {\bf Yan Yan} received the B.S. degree in Electrical Engineering from University of Electronic Science and Technology of China (UESTC), Chengdu, China, in 2004 and the Ph.D. degree in Information and Communication Engineering from Tsinghua University, Beijing, China, in 2009. He worked at Nokia Japan R\&D center (2009-2010) and Panasonic Singapore Lab (2011) as a research engineer and a project leader, respectively. He is currently an assistant professor in the School of Information Science and Technology at Xiamen University, Xiamen, China. His research interests include image recognition and machine learning.

\parpic{\includegraphics[width=1in,height=1.25in,clip]{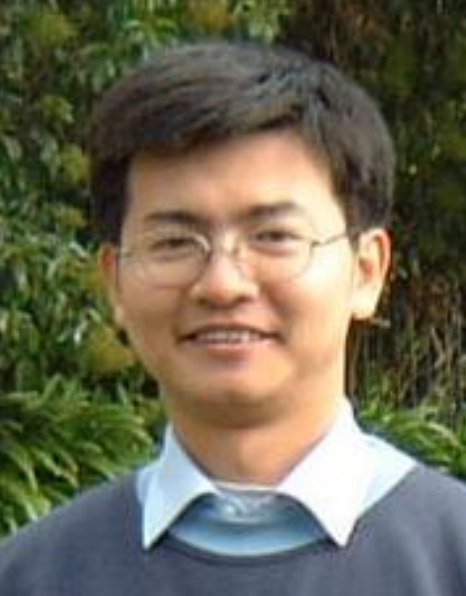}}
\noindent {\bf Hanzi Wang} is currently a Distinguished Professor and "Minjiang Scholar" at Xiamen University, China. He was a Senior Research Fellow (2008 - 2010) at the University of Adelaide, Australia; an Assistant Research Scientist (2007 - 2008) and a Postdoctoral Fellow (2006 - 2007) at the Johns Hopkins University; and a Research Fellow at Monash University, Australia (2004 - 2006). He received the Ph.D degree in Computer Vision from Monash University, Australia. He was awarded the Douglas Lampard Electrical Engineering Research Prize and Medal for the best PhD thesis in the Department. His research interests are concentrated on computer vision and pattern recognition including visual tracking, robust statistics, model fitting, object detection, video segmentation, and related fields. He has published more than 70 papers in major international journals and conferences including the IEEE Transactions on Pattern Analysis and Machine Intelligence, International Journal of Computer Vision, ICCV, CVPR, ECCV, NIPS, MICCAI, etc. He is an Associate Editor for IEEE Transactions on Circuits and Systems for Video Technology (T-CSVT) and he was a Guest Editor of Pattern Recognition Letters (September 2009). He is a Senior Member of the IEEE. He has served as a reviewer for more than 40 journals and conferences.

\end{document}